%% file: smws.tex
\documentclass[10pt,twocolumn,letterpaper]{article}

\usepackage{cvpr}
\usepackage{times}
\usepackage{epsfig}
\usepackage{graphicx}
\usepackage{amsmath}
\usepackage{amssymb}

\usepackage[utf8]{inputenc}
\usepackage[ngerman,english]{babel}
\usepackage{
    amsthm,
    booktabs,
    caption,
    dsfont,
    etoolbox,   %
    enumitem,   %
    mathtools,
    subcaption,
}
\usepackage{xcolor}

\usepackage{mathrsfs}

\usepackage[boxruled,ruled,vlined]{algorithm2e}
\DecMargin{5pt}
\definecolor{boxcolor}{rgb}{0.6, 0.81, 0.93}
\def\HiLi{\leavevmode\rlap{\hbox to \hsize{\color{boxcolor!50}\leaders\hrule height .8\baselineskip depth .5ex\hfill}}}
\DontPrintSemicolon
\SetKwInOut{KwInit}{Initialization}
\SetKwFunction{Merge}{merge}
\SetKwFunction{Connected}{connected}
\SetKwFunction{Mutex}{mutex}
\SetKwFunction{Addmutex}{addMutex}
\SetKwFunction{Class}{class}
\SetKwFunction{Assignclass}{assignLabel}
\SetKw{Not}{\textbf{not}}
\SetKw{And}{\textbf{and}}
\SetKw{Or}{\textbf{or}}
\SetKwComment{tcp}{$\quad \rhd$\ }{}
\SetCommentSty{textup}
\AtBeginEnvironment{algorithm}{\linespread{1.1}\selectfont}

\newlength\savewidth\newcommand\shline{\noalign{\global\savewidth\arrayrulewidth
  \global\arrayrulewidth 1pt}\hline\noalign{\global\arrayrulewidth\savewidth}}

\newcommand{\tss}[1]{\textsuperscript{#1}}

\makeatletter
\renewcommand\paragraph{\@startsection{paragraph}{4}{\z@}%
  {.1em \@plus1ex \@minus.1ex}%
  {-1em}%
  {\normalfont\normalsize\bfseries}}
\makeatother

\newtheorem{theorem}{Theorem}[section]

\newtheorem{lemma}[theorem]{Lemma}

\newcommand{\things}{\tss{Th}\xspace}
\newcommand{\stuff}{\tss{St}\xspace}

\usepackage[
  pagebackref=true,
  breaklinks=true,
  colorlinks,
  bookmarks=false
  ]{hyperref}

\usepackage[english,
            nameinlink
           ]{cleveref}

\crefname{algocf}{algorithm}{algorithms}
\Crefname{algocf}{Algorithm}{Algorithms}

\newcommand{\highest}[1]{\textbf{#1}}
\newcommand{\altw}{w}

\usepackage{pifont}

\cvprfinalcopy %

\newcommand{\customfootnotetext}[2]{{%
  \renewcommand{\thefootnote}{#1}%
  \footnotetext[0]{#2}}}%

\ifcvprfinal\fi
\begin{document}

\title{The Semantic Mutex Watershed for Efficient Bottom-Up Semantic Instance Segmentation}

\author{Steffen Wolf$^{1}$\textsuperscript{*} \qquad Yuyan Li$^{1}$\textsuperscript{*} \qquad Constantin Pape$^{2}$\\
        Alberto Bailoni$^{1}$ \qquad Anna Kreshuk$^{2}$ \qquad Fred A. Hamprecht$^{1}$\\[0.5cm]
    $^1$HCI/IWR, Heidelberg University, Germany \qquad $^2$EMBL Heidelberg, Germany
}

\maketitle

\input{chapters/abstract}
\customfootnotetext{*}{Equal contribution.}

\input{chapters/introduction}

\input{chapters/relatedwork}

\newpage
\input{chapters/theory}

\input{chapters/experiments}

\input{chapters/conclusion}

{\small
\bibliographystyle{ieee_fullname}
\bibliography{references}
}

\newpage
\renewcommand{\theequation}{A.\arabic{equation}}
\input{chapters/appendix}

\end{document}

%% file: chapters/abstract.tex
\begin{abstract}
\noindent Semantic instance segmentation is the task of simultaneously partitioning an image into distinct segments while associating each pixel with a class label.
In commonly used pipelines, segmentation and label assignment are solved separately since joint optimization is computationally expensive.
We propose a greedy algorithm for joint graph partitioning and labeling derived from the efficient Mutex Watershed partitioning algorithm~\cite{wolf2018mutex}.
It optimizes an objective function closely related to the Symmetric Multiway Cut objective and empirically shows efficient scaling behavior.
Due to the algorithm's efficiency it can operate directly on pixels without prior over-segmentation of the image into superpixels.
We evaluate the performance on the Cityscapes dataset (2D urban scenes) and on a 3D microscopy volume. %
In urban scenes, the proposed algorithm combined with current deep neural networks outperforms the strong baseline of `Panoptic Feature Pyramid Networks' by Kirillov \etal(2019).
In the 3D electron microscopy images, we show explicitly that our joint formulation outperforms a separate optimization of the partitioning and labeling problems.
\end{abstract}

%% file: chapters/introduction.tex
\section{Introduction}

Image segmentation literature distinguishes \emph{semantic segmentation} - associating each pixel with a class label - and \emph{instance segmentation}, i.e. detecting and segmenting individual objects while ignoring the background.
The joint task of simultaneously assigning a class label to each pixel and grouping pixels to instances has been addressed under different names, including semantic instance segmentation, scene parsing~\cite{tighe2015scene}, image parsing~\cite{tu2005image}, holistic scene understanding~\cite{yao2012describing} or instance-separating semantic segmentation~\cite{levinkov2017joint}.
Recently, a new metric and evaluation approach to such problems has been introduced under the name of \emph{panoptic segmentation}~\cite{kirillov2018panoptic}.

From a graph theory perspective, semantic instance segmentation corresponds to the simultaneous partitioning and labeling of a graph.
Most greedy graph partitioning algorithms are defined on graphs encoding attractive interactions only.
Clusters are then formed through agglomeration or division until a user-defined termination criterion is met (often a threshold or a desired number of clusters).
These algorithms perform pure instance segmentation. The semantic labels for the segmented instances need to be generated independently.

If repulsive - as well as attractive - forces are defined between the nodes of the graph, partitioning can be formulated as a Multicut problem~\cite{andres2011probabilistic}. In this formulation clusters emerge naturally without the need for a termination criterion.
Furthermore, the Multicut problem can be extended to include the labeling of the graph, delivering a semantic instance segmentation from a joint optimization of partitioning and labeling ~\cite{kroeger2014asymmetric}. The main drawback of this formulation is that the Multicut problem is NP-hard.

We propose to solve the joint partitioning and labeling problem by an efficient algorithm which we term Semantic Mutex Watershed (SMW), inspired by the Mutex Watershed~\cite{wolfmutex}. 
In more detail, in this contribution we:
\begin{itemize}[nosep]
   \item propose a fast algorithm for joint graph partitioning and labeling
   \item prove that the algorithm minimizes (exactly) an objective function closely related to the Symmetric Multiway Cut objective
   \item demonstrate competitive performance on natural and biological images. 
\end{itemize}

%% file: chapters/relatedwork.tex
\section{Related Work}
\paragraph{Semantic segmentation.}
State-of-the-art semantic segmentation algorithms are based on convolutional neural networks (CNNs) which are trained end-to-end.
The networks commonly follow the design principles of image classification networks (\eg~\cite{he2015deep,simonyan2014very,krizhevsky2017imagenet}), replacing the fully connected layers at the end with convolutional layers to form a fully convolutional network ~\cite{long2015fully}.
This architecture can be further extended to include encoder-decoder paths ~\cite{ronneberger2015unet}, dilated or atrous convolutions~\cite{yu2015multiscale,chen2016semantic} and pyramid pooling modules~\cite{chen2017rethinking,zhao2017pyramid}.

\paragraph{Instance segmentation.}
Many instance segmentation methods use a detection or a region proposal framework as their basis; object segmentation masks are then predicted inside region proposals.
A cascade of multiple networks is employed by~\cite{dai2016instanceaware}, each solving a specific subtask to find the instance labeling.
Mask-RCNN~\cite{he2017mask} builds on the bounding box prediction capabilities of Faster-RCNN~\cite{ren2015faster} to simultaneously produce masks and class predictions.
An extension of this method with an additional semantic segmentation branch has been proposed in~\cite{kirillov2019panoptic} as a single network for semantic instance segmentation.

In contrast to the region-based methods, proposal-free algorithms often start with a pixel-wise representation which is then clustered into instances~\cite{yu2018learning,kong2017recurrent,fathi2017semantic}. Alternatively, the distance transform of instance masks can be predicted and clustered by thresholding~\cite{bai2017deep}.

\paragraph{Graph-based segmentation.}
Graph-based methods, used independently or in combination with machine learning on pixels, form another popular basis for image segmentation algorithms ~\cite{felzenszwalb2004efficient}. In this case, the graph is built from pixels or superpixels of the image and the instance segmentation problem is formulated as graph partitioning. When the number of instances is not known in advance and repulsive interactions are present between the graph nodes, graph partitioning can in turn be formulated as a Multicut or correlation clustering problem~\cite{andres2011probabilistic}.
This NP-hard problem can be solved reasonably fast for small problem sizes with integer linear programming solvers~\cite{andres2012globally} or approximate algorithms~\cite{pape2017solving,beier2017multicut}.
A modified Multicut objective is introduced by ~\cite{wolfmutex} together with the Mutex Watershed - an efficient clustering algorithm for its optimization.%

The Multicut objective can be extended to solve a joint graph partitioning and labeling problem~\cite{kappes2011globally,kroeger2014asymmetric} for simultaneous instance and semantic segmentation.
In practice, the computational complexity of the joint problem only allows for approximate solutions ~\cite{levinkov2017joint}, possibly combined with reducing the problem size by over-segmentation into superpixels.
This formulation has been applied to natural images by~\cite{kirillov2017instancecut} and to biological images by~\cite{krasowski2018neuron}.

Similar to the semantic segmentation use case, CNNs can be used to predict pixel and superpixel affinities which serve as edge weights in the graph partitioning problem ~\cite{lee2017superhuman, maire2016affinity,liu2018affinity}.

%% file: chapters/theory.tex
\section{The Semantic Mutex Watershed}

The centerpiece of this paper, the Semantic Mutex Watershed algorithm, solves the semantic instance segmentation problem by jointly finding a graph partitioning and labeling.
In this section we present the graph-based formulation of the semantic instance segmentation problem and
define an objective function related to the Symmetric Multiway Cut problem~\cite{kroeger2014asymmetric}.
Then we introduce the Semantic Mutex Watershed algorithm and prove that it can optimize this objective efficiently.
Finally we show that the proposed objective constitutes a generalization of the Mutex Watershed Objective introduced in~\cite{wolfmutex}.

\subsection{Joint Partitioning and Labeling of Graphs} \label{sec:JointPartitioning}

Similar to instance segmentation algorithms, we build a graph of image pixels (voxels) or superpixels and formulate the semantic instance segmentation problem as joint partitioning and labeling of the graph.

\paragraph{Weighted graph with terminal nodes.}
For an undirected weighted graph $G = G(V, E, W)$ we refer to the nodes $V$ as \emph{internal nodes} and the edges $E$ as \emph{internal edges}.
We differentiate between \emph{attractive edges} $E^+$ and \emph{repulsive edges} $E^-$ that make up the internal edges $E = E^+ \cup E^-$.
Each edge $e \in E$ is associated with a real-valued positive weight $w_e \in W = W^+ \cup W^-$.
The weights encode the attraction and repulsion between the incident nodes of each edge.
A large \emph{attractive weight} $w_{uv} \in W^+$ encodes a high tendency for the nodes $u$ and $v$ to belong to the same partition element.
Equivalently, a large \emph{repulsive weight} $w_{uv} \in W^-$ indicates a strong inclination of $u$ and $v$ to belong to separate clusters.

\begin{figure}[t]
  (a) \vspace{-0.8cm} \\
  \includegraphics[width=\linewidth]{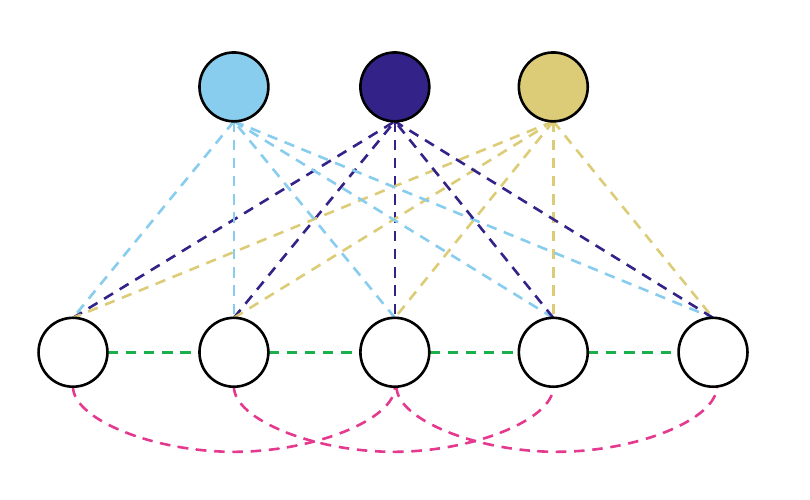} \\
  (b) \vspace{-0.8cm}\\
  \includegraphics[width=\linewidth]{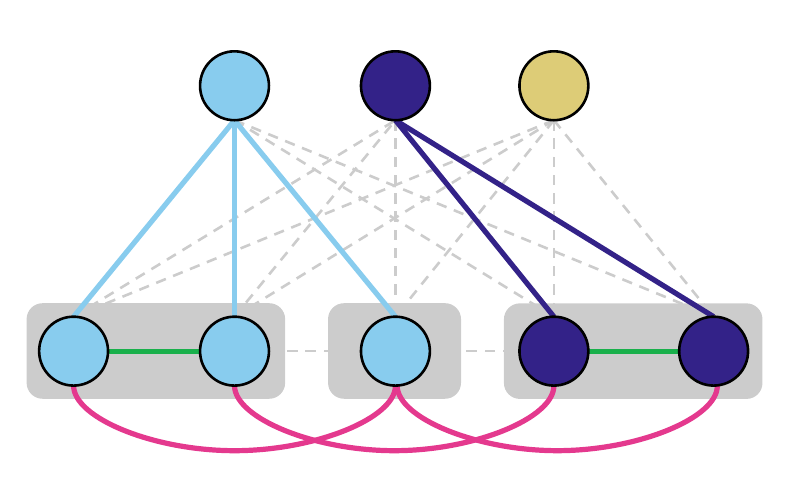}
  \captionof{figure}{(a) Example of an extended graph. Nodes on top are terminal nodes with each color representing a label class. The associated semantic edges are colored correspondingly. The internal nodes are on the bottom with attractive (green) and repulsive (red) edges between them. (b) Semantic instance segmentation. Edges that are part of the active set are shown in bold. Note that two adjacent nodes with the same label are not necessarily clustered together.}
  \label{fig:mmws_graph}
\end{figure}

Semantic instance segmentation can be achieved by clustering the internal nodes and assigning a semantic label $l \in \{l_0, ..., l_k\}$ to each cluster.
We extend $G$ by $k$ \emph{terminal} nodes $\{t_0, ..., t_k\} \in T$ where each $t_i$ is associated with a label $l_i$.
Every internal node $v \in V$ is connected to every $t$ by a weighted \emph{semantic edge} $e \in E^S$.
Here, a large \emph{semantic weight} $w_{ut} \in W^S \subseteq \mathbb{R}^+$ implies a strong association of internal node $u$ with the label of the terminal node $t$.
The extended graph thus becomes $G'(V', E', W')$ with $V' = V \cup T, \ E' = E \cup E^S$ and $ \ W' = W \cup W^S.$
\Cref{fig:mmws_graph}(a) shows a simple example of such an extended graph.

\begin{algorithm}[t]%
  \small
  \caption{\small The Semantic Mutex Watershed algorithm. The differences to the Mutex Watershed are marked in blue.} \label{alg:mmws}
  \KwIn{weighted graph $G'(V \cup T, E', W')$}
  \KwOut{clusters and labeling defined by $A$}
  \KwInit{$A = \emptyset$}
  \For{$(i, j) = e \in E'$ \textup{in descending order of} $w_e$}
  {
    \If{$e \in E^+$}
    {
      \If{\upshape \Not \Mutex{i, j} \\
      \HiLi \And \Not \Class{i} $\neq$ \Class{j}}
     {
       \Merge{i,j}: $A \gets A \cup e$\;
     }
    }
    \ElseIf{$e \in E^-$}
    {
      \If{\upshape \Not \Connected{i, j}}
      {
        \Addmutex{i, j}: $A \gets A \cup e$\;
      }
    }
    \HiLi\ElseIf{$e \in E^S$}
    {
      \HiLi\If{$\Class{i} = \emptyset$ \Or $\Class{i} = l_j$}
      {
        \HiLi{\Assignclass{i, j}: $A \leftarrow A \cup e$}\;
      }
    }
  }
  \Return{$A$}
\end{algorithm}

\newpage
\paragraph{Symmetric Multiway Cut.} 
In the Symmetric Multiway Cut an optimal semantic instance segmentation of such a graph is formulated as a constrained energy minimization/integer linear program (ILP)~\cite{kroeger2014asymmetric}:
\begin{align}
  \max_{a \in \{0,1\}^{|E'|}} \ \sum_{e \in E'} w_e^p \mathbf{a}_e \label{eq:objective}\\
  \sum_{(i, j) \in P \cap E^-} \mathbf{a}_{i j} + \sum_{(i, j) \in P\cap E^+} (1 - \mathbf{a}_{i j})   \geq (1 - \mathbf{a}_{u v}) \label{eq:objective_cycle_plus}\\ 
  \forall(u, v) \in E^+ ; \forall P \in \operatorname{Path}(u, v) \subseteq E  \notag\\ 
  \sum_{(i, j) \in P \cap E^-} \mathbf{a}_{i j} + \sum_{(i, j) \in P\cap E^+} (1 - \mathbf{a}_{i j}) \geq \mathbf{a}_{u v} \label{eq:objective_cycle_minus} \\ 
  \forall(u, v) \in E^- ; \forall P \in \operatorname{Path}(u, v) \subseteq E  \notag\\ 
  \sum_{t \in T} \mathbf{a}_{i t} = 1 \quad \forall i \in V \label{eq:objective_unique} \\
  (1 - \mathbf{a}_{u v})  \geq \mathbf{a}_{v t}-\mathbf{a}_{u t} \quad \forall(u, v) \in E^+ ; t \in T \label{eq:objective_consitent0} \\
  (1 - \mathbf{a}_{u v})  \geq \mathbf{a}_{u t}-\mathbf{a}_{v t} \quad \forall(u, v) \in E^+ ; t \in T \label{eq:objective_consitent1}
\end{align}
where $p=1$. The segmentation consistency is ensured by the cycle inequalities (\ref{eq:objective_cycle_plus}) and (\ref{eq:objective_cycle_minus}). 
\Cref{eq:objective_unique} enforces that every node is uniquely assigned to a terminal node. \Cref{eq:objective_consitent0,eq:objective_consitent1} ensure the consistency between partition labeling and semantic labeling.
A detailed discussion of the objective's properties will follow in \cref{sec:proof} and the relation to the objective in~\cite{kroeger2014asymmetric} is derived in \cref{app:rel_SMC}. Although this is \emph{in general} a hard optimization problem, we will show that for sufficiently large $p$ this objective can be solved exactly and efficiently by the algorithm introduced in the next section. 

\subsection{The Semantic Mutex Watershed Algorithm.}

We will now introduce a simple algorithm that greedily constructs a solution to \cref{eq:objective}. Although this will most likely not be an optimal solution to the NP-hard Symmetric Multiway Cut in general, we will show in \cref{sec:proof} that it becomes optimal when $p$ is large.

The clustering and label assignment of $G$ is described by a set of \emph{active} edges, which are chosen by the algorithm: $A \subseteq E'$ where $A \cap E^+$, $A \cap E^-$ and $A \cap E^S$ encode merges, mutual exclusions and label assignments, respectively. In order to restrict $A$ to a consistent partitioning and labeling we will make the following definitions:

We define two internal nodes $i,j \in V$ as connected if they are connected by active attractive edges, i.e.
\begin{equation}
  \Connected{i, j} \Leftrightarrow \exists \pi_{i \leadsto j} \subseteq A \cap E^+
  \label{eq:connected}
\end{equation}
Here $\pi_{i \leadsto j}$ denotes a path from node $i$ to node $j$.
We also define the mutual exclusion between two nodes as
\begin{equation}
  \Mutex{i, j} \Leftrightarrow \exists \, \pi_{i \leadsto j} \subseteq A \text{ with } |\pi \cap E^-| = 1.
  \label{eq:mutex}
\end{equation}
Two nodes are thus mutual exclusive if they are connected by a path from $i$ to $j$ with exactly one repulsive edge.
Furthermore, a label $l_j$ is assigned to a node $i$ if this node is connected to the corresponding terminal node $t_j$ by attractive and semantic edges:
\begin{equation}
  \Class{i} = l_j \Leftrightarrow \exists \, \pi_{i \leadsto t_j} \subseteq A \cap (E^+ \cup E^S).
  \label{eq:assignlabel}
\end{equation}
For unlabeled nodes we use the notation $\Class{i} = \emptyset$.

\paragraph{Algorithm.}

The Semantic Mutex Watershed algorithm is an extension of the Mutex Watershed algorithm introduced by~\cite{wolf2018mutex}.
It augments the partitioning of the latter with a consistent labeling.
The algorithm is shown in \cref{alg:mmws} with the additions to~\cite{wolf2018mutex} highlighted.
In the following we explain the syntax and procedure of the shown pseudocode.

All edges $E'$ are sorted in descending order of their weight and put in a priority queue.
While traversing the queue, the decision to add an edge to the set $A$ is made depending on the type of edge:

\noindent \emph{Attractive edges:} The edge is added if the incident nodes are not mutual exclusive and not labeled differently.

\noindent \emph{Repulsive edges:} The edge is added if the incident nodes are not connected.

\noindent \emph{Semantic edges:} The edge is added if the node is either unlabeled or already has the same label as the edge's terminal node.

Following these rules, the set of attractive edges in the final set $A \cap E^+$ form clusters in the graph, which are each connected to a single terminal node indicating the labeling.
\Cref{fig:mmws_graph}(b) shows a simple example of such an active set.

\paragraph{Efficient Implementation with Maximum-Spanning-Trees.}
The SMW is similar to the efficient Kruskal's maximum spanning tree algorithm~\cite{kruskal1956shortest} and can feasibly be applied to pixel-graphs of large images and even image volumes. Our implementation utilizes an efficient union-find data structure, mutex relations are realized/searched through a hash table. %

\paragraph{Mutex Watershed as Special Case.}
The Mutex Watershed algorithm is embedded in the Semantic Mutex Watershed as the special case when there are zero or one label ($|T| \in \{0, 1\}$).

\subsection{The Semantic Mutex Watershed Objective}\label{sec:proof}

In this section we prove that the Semantic Mutex Watershed Algorithm solves the ILP objective in \cref{eq:objective} for sufficiently large (\emph{dominant}) powers $p$.
To this end, we will extend the proof of~\cite{wolfmutex} by semantic edges.
First, we will review the definitions of \emph{dominant powers} and \emph{mutex constraints}. Second, we introduce an additional set of constraints acting on semantic edges and use it to define the Semantic Mutex Watershed Objective as a relaxed version of the Symmetric Multiway Cut.
Finally, we prove that the solution found by the SMW is indeed optimal.

\paragraph{Dominant Power.}
Let $\mathcal{G} = (V, E, W)$ be an edge-weighted graph, with unique weights $w_e \in \mathbb{R}^+_0,
\; \forall  e \in E$. We call $p \in \mathbb{R}^+$ a dominant power if:
\begin{equation}
    w_e^p > \sum_{s \in E,\; w_s < w_e} w_s^p \qquad \forall e \in E,
    \label{eq:pcondition}
\end{equation}
\noindent Note that there exists a dominant power for any finite set of edges, since for any $e \in E$ we can divide (\ref{eq:pcondition}) by $w_e^p$ and observe that the normalized weights $w_s^p/w_e^p$ (and any finite sum of these weights) converges to 0 when $p$ tends to infinity.

\paragraph{Semantic Mutex Watershed Constraints.}
To formalize the rules of the algorithm defined above, we first define special subsets of the active set $A$.
First, the set of all cycles containing exactly one repulsive edge is defined as
\begin{equation}
  \mathcal{C}_{1}(A) :=
  \left\{
    c \in \text{cycles}(\mathcal{G})
    \mid c \subseteq A \cap E
    \text{ and } |c \cap E^-| = 1
  \right\}.
  \label{eq:mutex_cycles}
\end{equation}
We define the \textbf{mutex constraint} as requiring $\mathcal{C}_1(A) = \emptyset$, which is exactly the rule that two mutual exclusive nodes must not be connected.

Furthermore, we define the set $\mathcal{P}(A)$ of all paths $\pi$ that connect two distinct terminal nodes through attractive and semantic edges:
\begin{equation}
  \mathcal{P}(A) := \{\, \pi_{t \leadsto t'} \subseteq A \cap (E^+ \cup E^S) \mid t,t' \in T\,\}
  \label{eq:mw_paths}
\end{equation}
The algorithm must never connect two terminal nodes through such a path, thus we define the \textbf{label constraint} $\mathcal{P}(A) = \emptyset$.
This ensures the consistency between the partitioning and labeling.
The mutex and label constraint are necessary but not sufficient to fulfill the linear constraints in \cref{eq:objective_cycle_plus,eq:objective_cycle_minus,eq:objective_unique,eq:objective_consitent0,eq:objective_consitent1} (the formal derivation can be found in \cref{app:label_contraints}).

\begin{lemma}[\bf Optimality of the Semantic Mutex Watershed]
Let $G = G(V', E', W') = G(V \cup T, E \cup E^S, W \cup W^S)$ be an edge-weighted graph extended by terminal nodes $T$, with unique weights $w_e \in \mathbb{R}^+_0$  and $p \in \mathbb{R}^+$ a dominant power.
Then the Semantic Mutex Watershed \Cref{alg:mmws} finds the optimal solution to the integer linear program
\begin{align}
  \max_{a \in \{0,1\}^{|E'|}}  & \sum_{e \in E'} \ w_e^p a_e \label{eq:mmws_energy}\\
  \textup{s.t.} \quad
    & \mathcal{C}_{1}(A) = \emptyset, \label{eq:mmws_cycle_1}\\
    & \mathcal{P}(A) = \emptyset, \label{eq:mmws_label}\\
 \textup{with} \quad
    & A := \{\, e \in E \mid a_e = 1 \,\}. \label{eq:mmws_activeset}
\end{align} \label{def:mmws_objective}
\vspace{-0.5cm}
\end{lemma} \vspace{-0.5cm}

\begin{proof}
 ~\cite{wolfmutex} show that for $T = \emptyset$ the SMW finds the optimal solution because it enjoys the properties \emph{greedy choice} and \emph{optimal substructure}.
Their proof of optimal substructure does not rely on the specific constraints in the ILP.
Thus it can also be applied with the additional constraint in \cref{eq:mmws_label}, giving the ILP \cref{eq:mmws_energy,eq:mmws_cycle_1,eq:mmws_label,eq:mmws_activeset} optimal substructure.

In every iteration the SMW adds the feasible edge $e$ with the largest weight to the active set.
Due to the dominant power, its energy contribution is larger than for any combination of edges $e'$ with $w_e' < w_e$.
Thus, SMW has the greedy choice property~\cite{cormen2009introduction}.
It follows by induction that the SMW algorithm finds the globally optimal solution to the SMW objective.

\end{proof}
We can now finally observe that the SMW algorithm always yields a consistent graph partitioning and labeling which fulfills the Symmetric Multiway Cut constraints. Thus, the Semantic Mutex Watershed algorithm returns an optimal solution of \cref{eq:objective,eq:objective_cycle_plus,eq:objective_cycle_minus,eq:objective_unique,eq:objective_consitent0,eq:objective_consitent1} if $p$ is set to a dominant power. In particular, if $p=1$ is dominant then the SMW solution is also an optimal solution to the Symmetric Multiway Cut.

%% file: chapters/experiments.tex
\section{Experiments}\label{sec:experiments}

\newcommand{\incpr}[1]{\includegraphics[trim=100 200 100 100, clip, width=.23\linewidth]{images/city/images/#1.jpg}}
\newcommand{\incprpred}[1]{\includegraphics[trim=200 400 200 200, clip, width=.23\linewidth]{images/city/prediction/#1.png}}
\newcommand{\incprgt}[1]{\includegraphics[trim=200 400 200 200, clip, width=.23\linewidth]{images/city/gt/#1.png}}
\begin{figure*}[t]
\begin{subfigure}{1.0\linewidth}
  \centering
  \rotatebox{90}{\begin{minipage}{.0558\linewidth}\center\tiny image\end{minipage}}
  \incpr{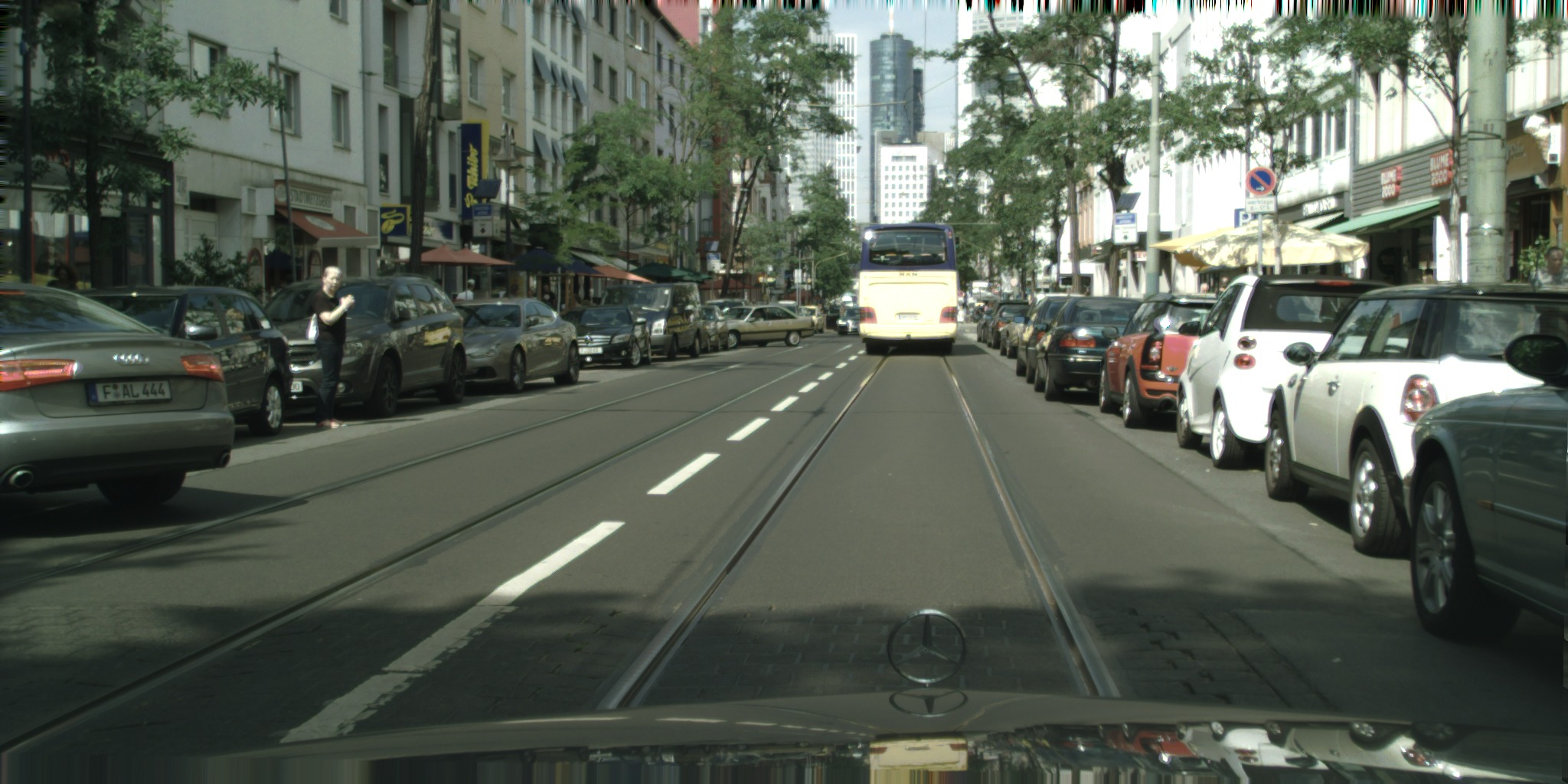}
  \incpr{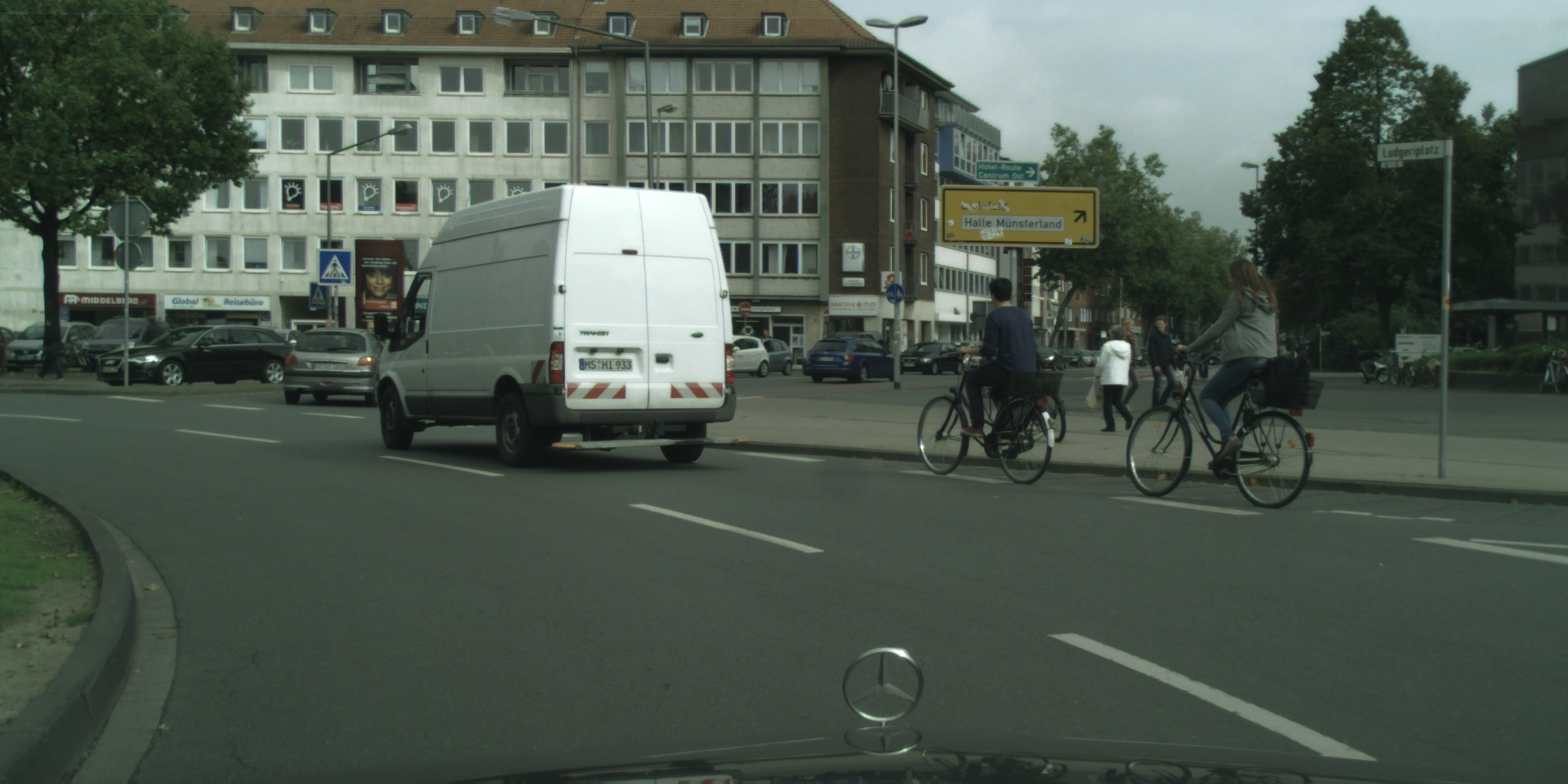}
  \incpr{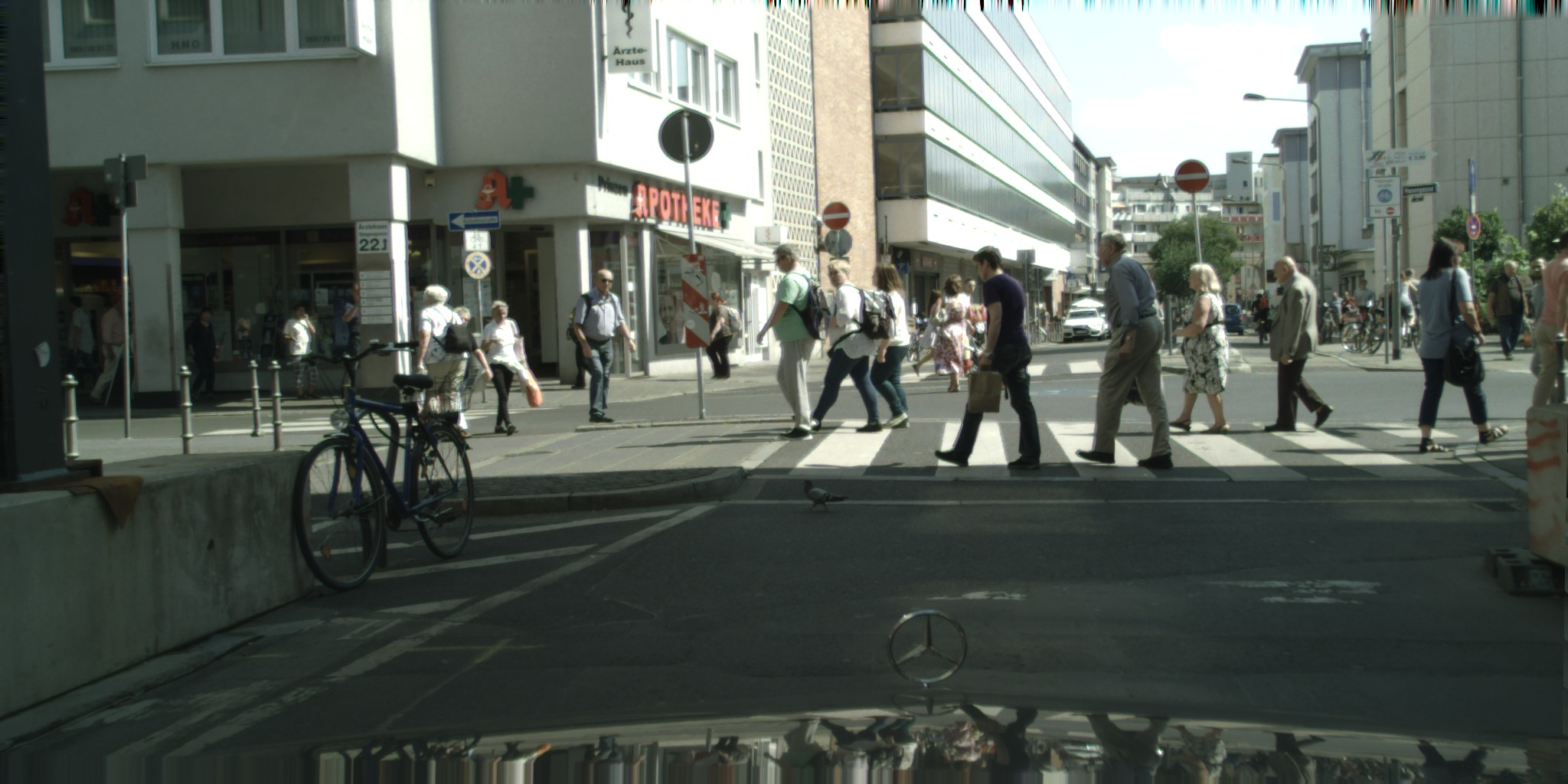}
  \includegraphics[width=.18\linewidth]{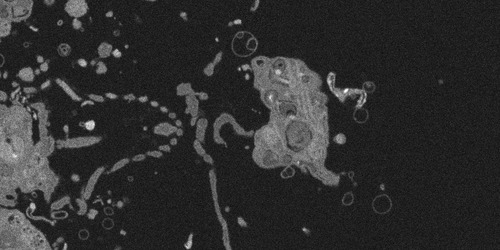}
\end{subfigure}\\[1pt]
\begin{subfigure}{1.0\linewidth}
  \centering
  \rotatebox{90}{\begin{minipage}{.0558\linewidth}\center\tiny groundtruth\end{minipage}}
  \incprgt{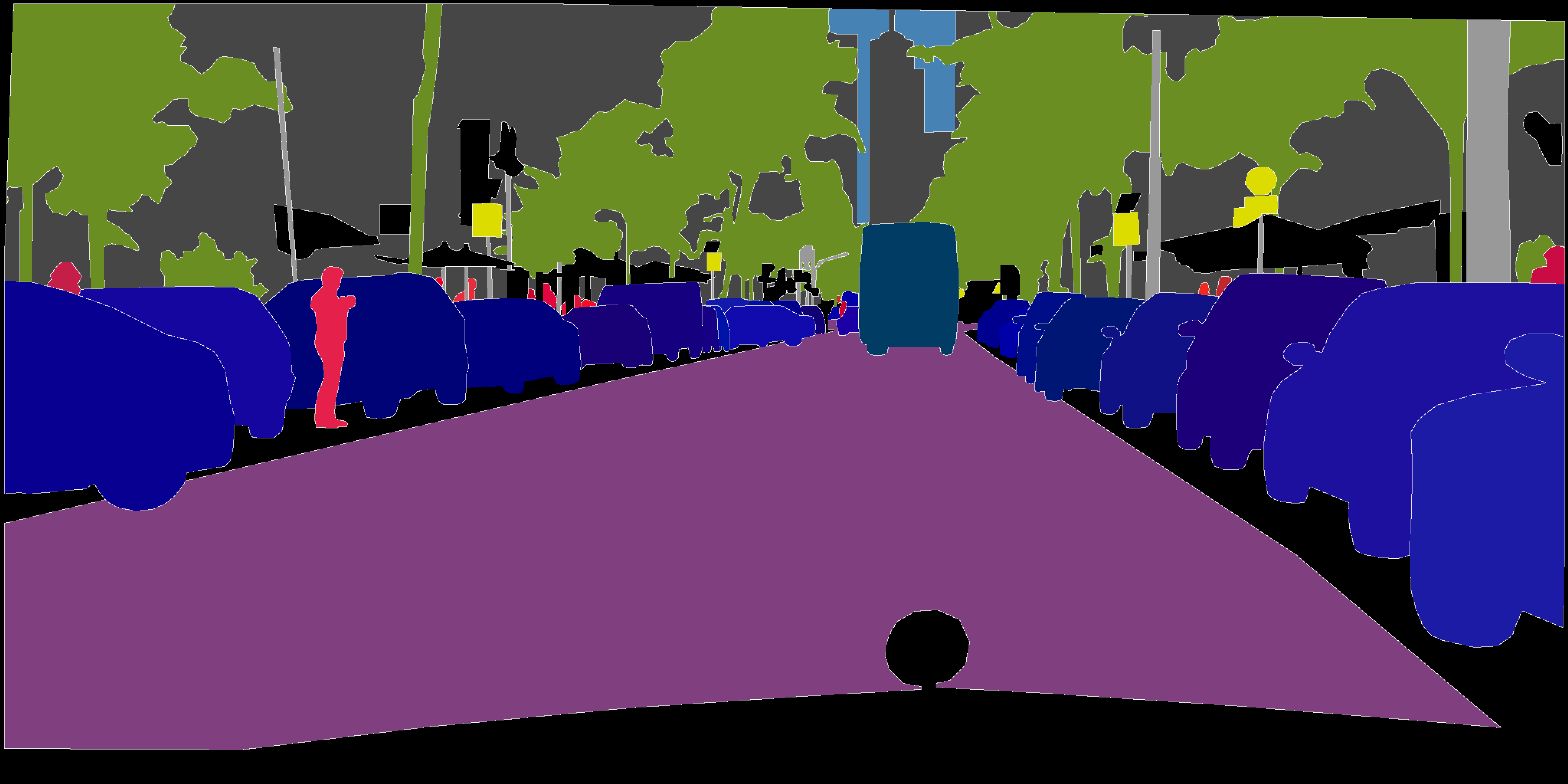}
  \incprgt{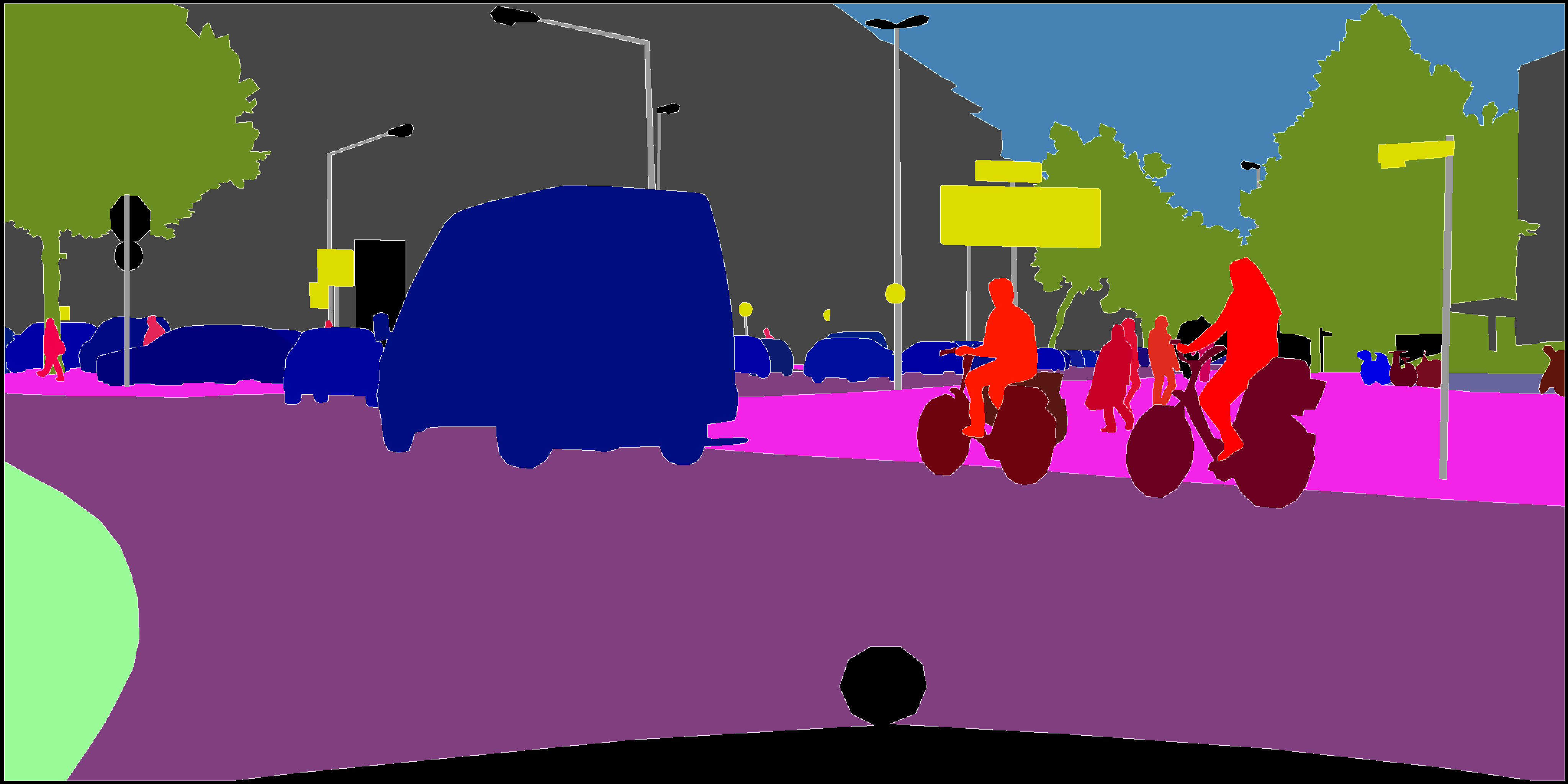}
  \incprgt{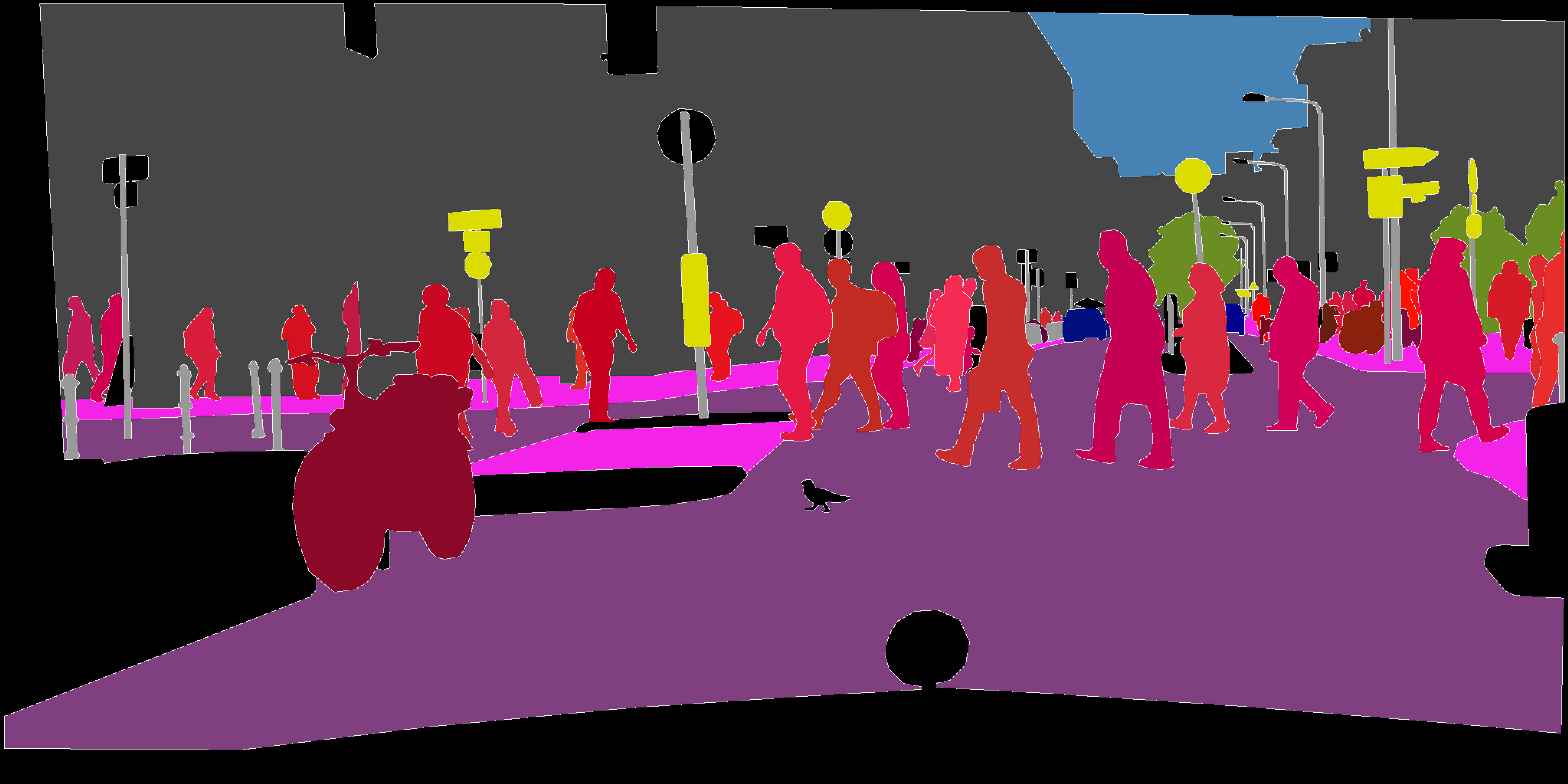}
  \includegraphics[width=.18\linewidth]{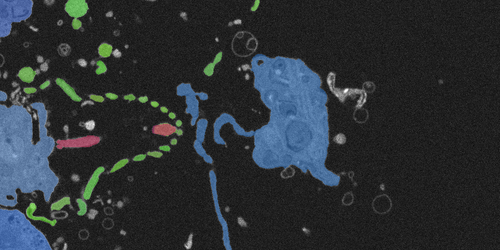}
\end{subfigure}\\[1pt]
\begin{subfigure}{1.0\linewidth}
  \centering
  \rotatebox{90}{\begin{minipage}{.0558\linewidth}\center\tiny prediction\end{minipage}}
  \incprpred{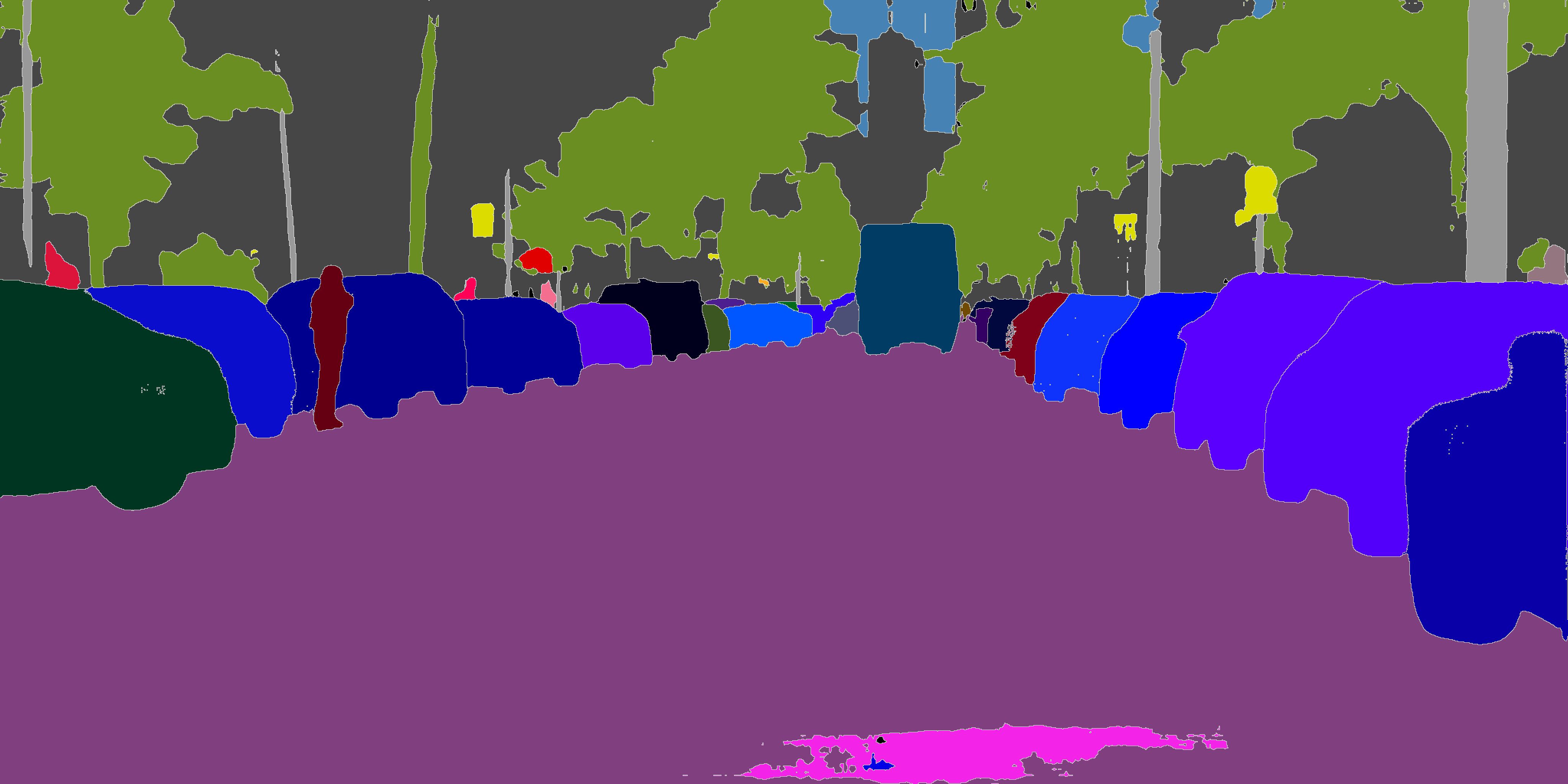}
  \incprpred{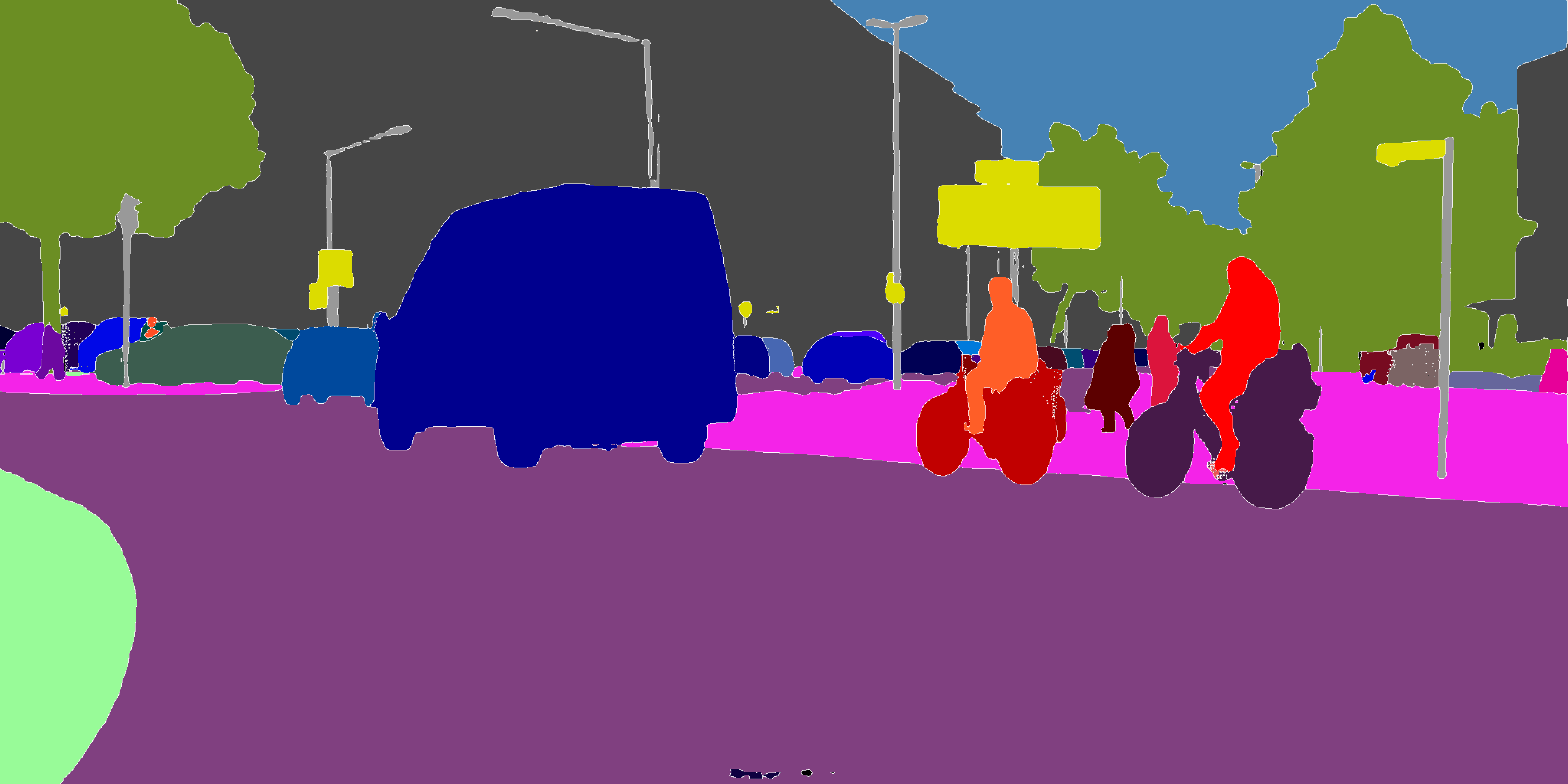}
  \incprpred{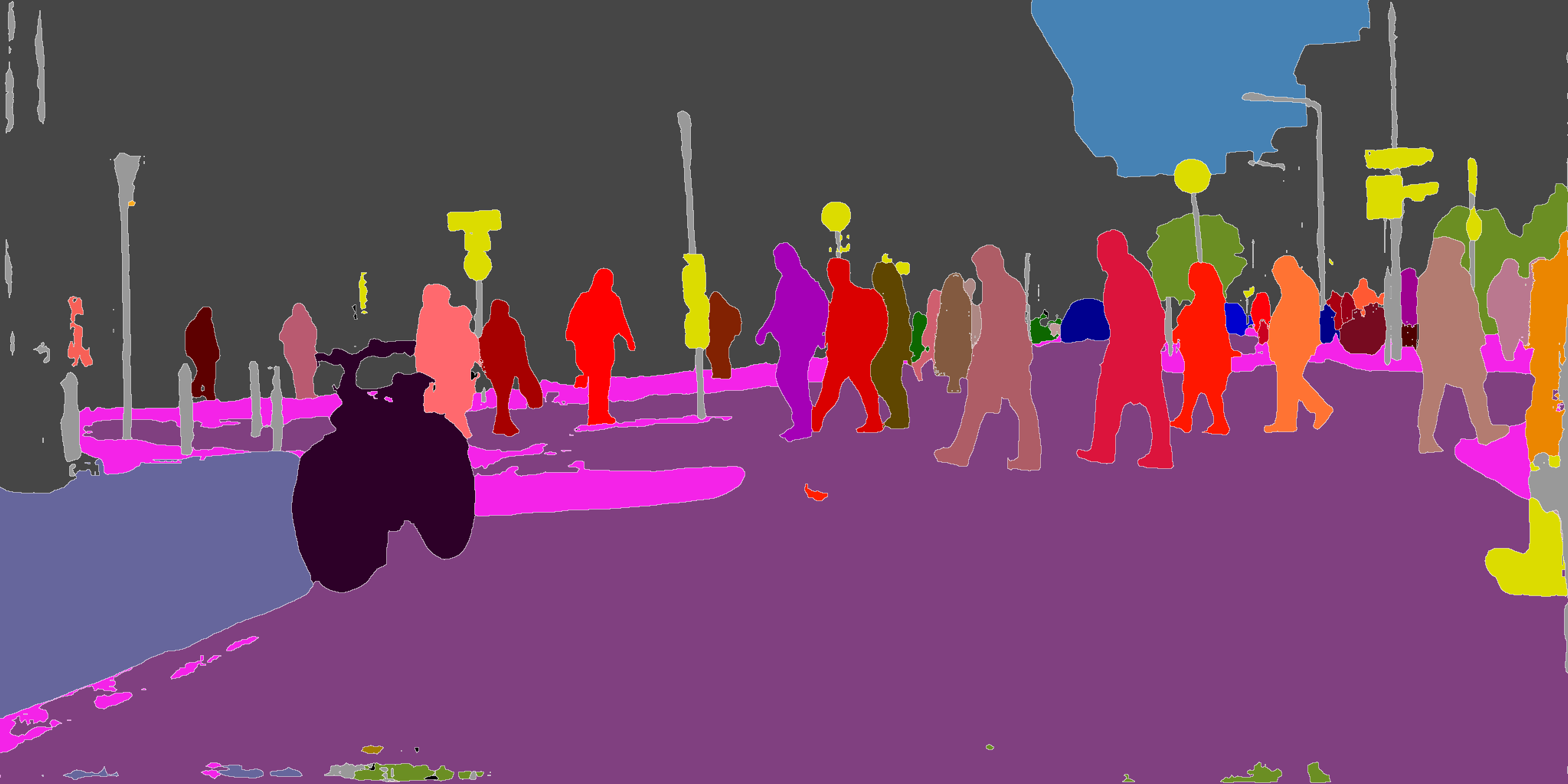}
  \includegraphics[width=.18\linewidth]{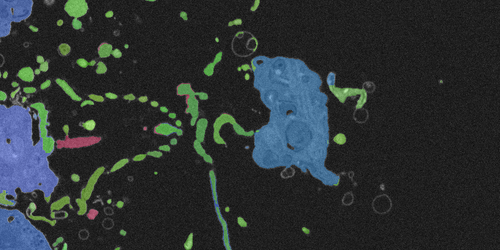}
\end{subfigure}
\vspace{2mm}
\caption{Semantic instance segmentation. \emph{First three columns:} Results on Cityscapes using semantic unaries (Deeplab 3+ network) and affinities derived from Mask-RCNN foreground probability. Colors indicate predicted semantic classes with variations for separate instances. \emph{Rightmost column:} Results for the sponge dataset. Cell-bodies are colored in blue, microvilli in green and flagella in red.}
\label{fig:examples}
\vspace{-2mm}
\end{figure*}

We will now demonstrate how to apply the SMW algorithm to semantic instance segmentation of 2D and 3D images.
We start from showing how existing CNNs can be used as graph weight estimators and compare different sources of edge weights on the Cityscapes dataset.
Additionally, we apply the SMW algorithm to a 3D electron microscopy volume  and demonstrate its efficiency and scalability.

\subsection{Affinity Generation with Neural Networks}

The only input to the SMW are the graph weights; it does not require any hyperparamters such as thresholds.
Consequently, its segmentation quality relies on good estimates of the graph weights $W' = W \cup W^S$. In this section we present how state-of-the-art CNNs can be used as sources for these weights.

\paragraph{Affinity Learning.}
Affinities are commonly used in instance segmentation; many modern algorithms train CNNs to directly predict pixel affinities.
A universal approach is to employ a stencil pattern that describes for each pixel which neighbours to consider for the affinity computation.
Regularly spaced, multi-scale stencil patterns are widely used for natural images~\cite{maire2016affinity,liu2018affinity} and bio-medical data~\cite{wolf2018mutex,lee2017superhuman}.

The predicted affinities are usually in the interval $[0,1]$ and can be interpreted as pseudo-probabilities. We use these affinities directly as weights for the attractive edges and invert them to get the repulsive edge weights.

\paragraph{Mask-RCNN}
 produces overlapping masks that have to be resolved for a consistent panoptic segmentation. We achieve this with the SMW by deriving affinities from the foreground probabilities of each mask.
A straightforward approach is to compute the (attractive) affinity $a(i,j)$ of two pixels as their joint foreground probability, weighted by the classification score $s$: $a(i,j) = s \, p(i) \, p(j)$.

We find that sparse repulsive edges work well in practice, as they lead to faster inference and reduced over-segmentation on the instance boundaries.
+For this reason, we sample random points from all pairs of masks and add (repulsive) edges with weight proportional to a soft intersection over union of two masks $m$ and $n$:
\begin{equation}
    w_{nm}= 1-\frac{\sum_{q \in V} p_{m}(q) p_{n}(q)}{\sum_{q \in V} \max{(p_{m}(q),\,\; p_{n}(q))}}.
    \label{eq:softiou}
\end{equation}

\paragraph{Semantic Segmentation CNNs.}
State of the art CNNs~\cite{chen2018encoderdecoder,zhao2017pyramid} achieve high quality results on semantic segmentation tasks.
The output of the last softmax layer usually used in these networks can be interpreted as the normalized probability of each pixel belonging to each class.
Thus, we can use these predictions directly as semantic weights $W^S$.

Additionally, we derive affinities from the stuff class probabilities; we treat each stuff class separately and again compute the affinity of two pixels as their joint probability of being in each stuff class $c$, i.e.: $a_c(i,j) = p_\text{c}(i) \, p_\text{c}(j)$. This cannot be done for thing classes since they can have multiple instances.

\subsection{Panoptic Segmentation on Cityscapes}

We apply the SMW on the challenging task of panoptic segmentation on the Cityscapes dataset~\cite{cordts2016cityscapes}.
We illustrate how the different sources of affinities can be used and combined and show their different strengths and weaknesses.

\paragraph{Dataset.}
The Cityscapes dataset consists of urban street scene images taken from a driver's perspective.
It has 5k densely annotated images separated into train (2975), val (500) and test (1525) set.
Since there is no public evaluation server for panoptic segmentation on the test set, we report all results on the validation set.
There are 19 classes with 11 stuff classes and 8 thing classes.

\paragraph{Implementation Details.}
We employ and combine multiple sources of graph weights to build the SMW graph. 
We train a Deeplab 3+~\cite{chen2018encoderdecoder} network for semantic edge weight and affinities prediction following~\cite{liu2018affinity}. We employ the Mask-RCNN~\cite{he2017mask} implementation provided by~\cite{massa2018mrcnn} and train a model on Cityscapes following~\cite{he2017mask}'s training configuration. Further implementation details can be found in \cref{app:city_imp_appendix}.
\begin{table*}
\begin{minipage}[b]{.5\textwidth}
\centering
\tabcolsep=0.08cm
\begin{tabular}{@{}cc|cc|ccc|llllll@{}}
\multicolumn{2}{c|}{MRCNN\cite{he2017mask}} & \multicolumn{2}{c|}{GMIS\cite{liu2018affinity} } & \multicolumn{3}{c|}{Deeplab\cite{chen2018encoderdecoder}}  &\multicolumn{3}{c}{{\bf Cityscapes}} \\
    att & rep & att & rep & att & rep & sem & PQ & PQ\things & PQ\stuff\\ \shline
\ding{51}&\ding{51}&&&&&\ding{51} & 59.3& 50.6& 65.7 \\ %
&\ding{51}&\ding{51}&&&&\ding{51} & 58.6& 48.8& 65.7 \\
&&\ding{51}&\ding{51}&&&\ding{51} & 56.1& 42.8& 65.7 \\ %
\ding{51}&\ding{51}&\ding{51}&\ding{51}&\ding{51}&\ding{51}&\ding{51} & 48.7& 38.7& 55.9 \\ %
&&\ding{51}&\ding{51}&\ding{51}&\ding{51}&\ding{51} & 47.3& 35.5& 55.9 \\ %
&&&&\ding{51}&\ding{51}&\ding{51} & 46.3& 33.1& 56.0 \\ %
\rule{0pt}{0pt}    
\end{tabular}
\caption{Panoptic segmentation quality PQ of the SMW on top of diverse sources of graph weights.}
\label{tab:Abalation}
\end{minipage}\hfill
\begin{minipage}[b]{.5\textwidth}
\centering
\tabcolsep=0.09cm
\begin{tabular}{@{}llll@{}}
{\bf Cityscapes} & PQ & PQ\things &  PQ\stuff \\ \shline
SMW & \highest{59.3}& 50.6& \highest{65.7} \\
PFPN\cite{kirillov2019panoptic} & 58.1 & \highest{52.0}& 62.5 \\
DIN\cite{arnab2017pixelwise} & 53.8 & 42.5& 62.1 \\
\rule{0pt}{3ex}    
{\bf Sponge} & PQ & PQ\things & PQ\stuff \\ \shline
SMW & \highest{51.6}& \highest{62.1}& 20.0 \\ %
MWS-MAX & 48.1& 56.2& \highest{23.8}  \\ %
CC$_{\text{sem}}$ & 43.4& 55.6& 06.7  \\ %
CC$_{\text{aff}}$ & 24.3& 27.7& 13.9  \\ %
\rule{0pt}{0.0cm}
\end{tabular}
\caption{Comparison to other segmentation strategies.}%
\label{tab:city_full}
\end{minipage}

\end{table*}

\paragraph{Study of Affinity Sources.}

We evaluate the semantic instance segmentation performance of the SMW in terms of the ``panoptic'' metric using different combinations of the graph weight sources discussed above.
In \cref{tab:Abalation} we compare the PQ metric on the Cityscapes dataset.

The best performance can be achieved with a combination of Mask-RCNN affinities and Deeplab 3+ for semantic predictions outperforming the strong baseline of~\cite{kirillov2019panoptic} listed in \cref{tab:city_full} and shown in \cref{fig:examples} and the supplementary \cref{fig:examples_apdx}.
Through observations on the images, we find that Mask-RCNN affinities are more reliable in detecting small objects as well as in connecting fragmented instances. Note that PQ mostly measures detection quality which is then weighted by the segmentation quality of the found instances, hence the detection strength of the Mask-RCNN shines through.

We observe that using all sources together leads to a performance drop of 10 percentage points below the best result.
We believe this is due to the greedy nature of the SMW which selects the strongest of all provided edges. This example demonstrates how important it is to carefully select/train the algorithm input.

\subsection{Semantic Instance Segmentation of 3D EM Volumes}

Semantic instance segmentation is an important task in bio-medical image analysis where classes naturally arise through cellular structure. We use a 3D EM image dataset to compare the SMW to algorithms that separately optimize instance segmentation and semantic class assignment.

\paragraph{Dataset.}
The data-set consists of two FIBSEM volumes of a sponge choanocye chamber. The data was acquired in \cite{musser2019profiling} to investigate proto-neural cells in sponges using the segmentation approach introduced in \cite{pape2019leveraging}. These cells filter nutrients from water by creating a flow with the beating of a flagellum and absorbing the nutrients through microvilli that surround the flagellum in a collar~\cite{langenbruch1987canal} (see \cref{fig:examples}).
In order to investigate this process in detail, a precise semantic instance segmentation of the cell-bodies, flagella and microvilli is needed. The dataset consists of three EM image volumes of size $96 \times 896 \times 896$ pixel ($2 \times 18 \times 18$ $\mu$m).

\paragraph{Implementation Details.}
We predict affinities with two separate 3D U-Nets~\cite{cciccek20163d} to derive graph edge weights and semantic class probabilities respectively.
We adopt the training procedure of~\cite{wolf2018mutex} which uses the Dice Coefficient as the loss function.
We use two volumes for training and one for testing.

We also implement baseline approaches which start from the same network predictions, but do not perform joint labeling and partitioning.
First, we compare to instance segmentation with the Mutex Watershed, followed by assigning instances the semantic label of the strongest semantic edge (MWS-MAX).
In addition, we compute connected components of the semantic predictions ($CC_\text{{sem}}$) and short-range affinities ($CC_{\text{aff}}$).

\paragraph{Results.}
The PQ values in \cref{tab:city_full} show that the SMW outperforms the baselines approaches that separately optimize instance segmentation and semantic class assignment.
An additional analysis can be found in the appendix \cref{fig:rt_apdx}, where we measure the runtimes for different volume sizes and observe almost linear scaling behavior.

%% file: chapters/conclusion.tex
\section{Conclusion}
We have introduced a new method for joint partitioning and labeling of weighted graphs as a generalization of the Mutex Watershed algorithm.
We have shown that it optimally solves an objective function closely related to the objective of the Symmetric Multiway Cut problem.
Our experiments demonstrate that SMW with graph edge weights predicted by convolutional neural networks outperform strong baselines on natural and biological images. Any improvement in the CNN performance will translate directly to an improvement of the SMW results. However, we also observe that the extreme value selection used by the SMW to assign edges to the active set can lead to sub-optimal performance when diverse edge weights sources are combined. 
Empirically, the algorithm scales almost linearly with the number of graph edges $N$ making it applicable to large images and volumes without prior over-segmentation into superpixels.
The source code will be made available upon publication.

%% file: chapters/appendix.tex
\appendix

\section{Symmetric Multiway Cut in Related Literature}\label{app:rel_SMC}

We will now relate the Symmetric Multiway Cut definition in \cref{eq:objective,eq:objective_cycle_plus,eq:objective_cycle_minus,eq:objective_unique,eq:objective_consitent0,eq:objective_consitent1} with the the objective given in~\cite{kroeger2014asymmetric}. In contrast to this work Kroeger \etal~\cite{kroeger2014asymmetric} do not split the set of edges in to attractive and repulsive edges. Instead they model repulsion with negative weights and formulate the SMWC as the following constrained energy minimization/integer linear program (ILP):
\begin{equation}
 \min_{y \in \{0,1\}^{|E'|}} \ \left( \sum_{e \in E^S} \altw_e (1 - y_e) + \sum_{e \in E} \altw_e y_e \right) \quad 
  \text {s.t.} \
 y \in \operatorname{SMWC}_{G^{\prime}}
\label{eq:amwc_energy}
\end{equation}
The variables $y_e \in \{0, 1\}$ are indicators for cuts in the graph, i.e. when $y_e = 1$ the edge $e$ is cut, and $\operatorname{SMWC}_G'$ is the polytope of consistent solutions defined by linear constraints:
\begin{align}
 \sum_{e \in c \setminus \{ e^- \}} y_e \geq y_{e^-} \quad & \forall e^- \in c \ \forall c \in \textrm{Cycles in } G \label{eq:amwc_cycle}\\
 \sum_{t \in T} y_{it} = |T| - 1 \quad & \forall i \in V \label{eq:amwc_label}\\
 y_{uv} \geq y_{ut} - y_{vt} \quad & \forall (u, v) \in E; t \in T \label{eq:amwc_mw_0}\\
 y_{uv} \geq y_{vt} - y_{ut} \quad & \forall (u, v) \in E; t \in T \label{eq:amwc_mw_1}
\end{align}

The cycle constraints (\ref{eq:amwc_cycle}) form the so called Multicut polytope~\cite{andres2011probabilistic}; they forbid dangling edges thus all non-cut internal edges form clusters on the graph $G$.
\Cref{eq:amwc_label} ensures that each internal node is connected to exactly one terminal node.
Finally, (\ref{eq:amwc_mw_0}) and (\ref{eq:amwc_mw_1}) are cycle constraints on all cycles with one terminal node; they enforce that an internal edge is always cut when its incident nodes are connected to different terminal nodes.
This ensures that the resulting partitioning and labeling is always consistent.
Note that an edge between two nodes connected to the same terminal is allowed to be cut, so two instances of the same class may touch.

We will now transform the objective given in \cref{eq:amwc_energy} and introduce an additional parameter $p$. Instead of finding a small-weight set to cut from the graph, we try to find a large-weight set $A \subseteq E$ to keep in the graph.

First, we split the internal edges into repulsive ($E^- := \{\, e \in E \mid \altw_e < 0 \,\}$) and attractive edges ($E^+ := \{\, e \in E \mid \altw_e \geq 0\}$) so the energy function becomes
\begin{equation}
\sum_{e \in E^S} \altw_e^p (1 - y_e) + \sum_{e \in E^+} \altw_e^p y_e  - \sum_{e \in E^-} |\altw_e|^p y_e.
\label{eq:transformed_energy}
\end{equation}
For $\boldsymbol p \boldsymbol = \boldsymbol1$ the ILP corresponds to the Symmetric Multiway Cut.
Subtracting the constant sum of all positive edge weights, using $\altw_e \leq 0 \ \forall e \in E^S$ 
yields
\begin{equation}
- \sum_{e \in E^S} |\altw_e|^p (1 - y_e) - \sum_{e \in E^+} \altw_e^p (1 - y_e)  - \sum_{e \in E^-} |\altw_e|^p y_e.
\end{equation}
Finally,
by substituting
\begin{equation}
a_e  =
\begin{cases}
y_e     & \textrm{if } e \in E^- \\
1 - y_e & \textrm{if } e \in E^+ \cup E^S
\end{cases}
\end{equation}
we obtain the equivalent objective
\begin{equation}
\begin{split}
  \max_{a \in \{0,1\}^{|E'|}} \ \sum_{e \in E'} |w_e|^p a_e \\
  \text{s.t.} \
   a \in \operatorname{SMWC}^a_{G'}.
\end{split}
\end{equation}
Here, $\operatorname{SMWC}^a_{G'}$ is the polytope formed by \cref{eq:objective_cycle_plus,eq:objective_cycle_minus,eq:objective_unique,eq:objective_consitent0,eq:objective_consitent1}. Since all weights are positive in the SMW graph, the absolute value is omitted in \cref{eq:objective}.

\section{Mutex and Label Constraints}\label{app:label_contraints}

We will now formality derive that the constraints in \cref{eq:mmws_cycle_1,eq:mmws_label} are necessary for the Symmetric Multiway Cut constraints \cref{eq:amwc_cycle,eq:amwc_label,eq:amwc_mw_0,eq:amwc_mw_1}. 
Wolf \etal~\cite{wolfmutex} show that the \cref{eq:mmws_cycle_1} is necessary for the multicut constraints \cref{eq:amwc_cycle}.
Therefore, it is left to show that
\begin{equation}
  \begin{rcases*}
    \sum_{t \in T} y_{ut} = |T| - 1 \quad \forall u \in V \\
    y_{uv} \geq y_{ut} - y_{vt} \quad \forall (u, v) \in E; t \in T \\
    y_{uv} \geq y_{vt} - y_{ut} \quad \forall (u, v) \in E; t \in T
  \end{rcases*}
  \Rightarrow \mathcal{P}(A) = \emptyset.
  \label{eq:amwc_paths}
\end{equation}
The right-hand side is the label constraint which will be shown to be a subset of the constraints formed by \cref{eq:amwc_label,eq:amwc_mw_0,eq:amwc_mw_1} (here on the left).

First we show by contradiction and using \cref{eq:amwc_label} that an internal node $v \in V$ can only be connected to a single terminal $t \in T$: Assume that there is a $t^* \neq t$ which is also connected to $v$; then we have $y_{ut} = 0$ and $y_{ut^*} = 0$. Now rewrite \cref{eq:amwc_label} and insert these two variables so that we get the contradiction
\begin{align}
  1 = \sum_{t' \in T} (1 - y_{ut'}) = \sum_{t' \in T \setminus \{t, t^*\}} (1 - y_{ut'}) + 1 + 1 \geq 2,
\end{align}

We further show that two connected nodes $u$ and $v$ are always connected to the same terminal node $t$. Without losing generality we assume $u$ and $v$ are connected and $u$ and $t$ are connected, i.e. $y_{uv} = 0$ and $y_{ut} = 0$. Then \cref{eq:amwc_mw_0,eq:amwc_mw_1} give us
\begin{equation}
  \begin{rcases*}
    0 \geq 0 - y_{vt}\\
    0 \geq y_{vt} - 0
  \end{rcases*}
  \Rightarrow y_{vt} = 0
\end{equation}

Finally, we can prove \cref{eq:amwc_paths}: any path starting from $t$ begins with an edge $(t,v)$ to some node $u$; all nodes connected to $u$ (and $u$ itself) are connected to $t$ and no other terminal node. Therefore there can not be any path from $t$ to another terminal node $t'$ satisfying the label constraint $\mathcal{P}(A) = \emptyset$.

\section{Additional Details of the Cityscapes Experiments}
\subsection{Implementation Details}\label{app:city_imp_appendix}
We use the class probabilities from a Deeplab 3+~\cite{chen2018encoderdecoder} as semantic edge weights. We use a trained model provided by Tensorflow.
employ the Mask-RCNN~\cite{he2017mask} implementation provided by~\cite{massa2018mrcnn} and trained a model on Cityscapes following~\cite{he2017mask}'s training configuration.
The graph weights are derived as explained above. We derive graph weights for different offsets: for attractive edges we use (1) 8-neighbourhood with distances of \{1, 2, 4\} pixels, (2) random pairs inside each bounding box. For repulsive edges we sample 5 random pixel pairs for each mask and compute the soft IOU (\cref{eq:softiou}).
\cite{liu2018affinity} trained a Deeplab 3+ to predict affinities for their graph-clustering algorithm. They kindly provided their trained models allowing us to use the same affinities. Since their clustering utilizes a threshold, we treat the threshold as the splitting point between attractive and repulsive edge weights; affinities below the threshold are inverted and scaled to [0, 1].
In addition to the model by GMIS that is trained on scaled bounding boxes, we train a Deeplab3+ for affinity predictions on the full images. Because~\cite{liu2018affinity} only tackle instance segmentation, their model does not predict affinities for stuff classes.
We train the network with Sorensen Dice Loss and the same stencil pattern as~\cite{liu2018affinity}.
The training protocol follows the settings in~\cite{chen2018encoderdecoder}, using a batch size of 12 and 70k training iterations.
We do not employ any test time augmentations.

\subsection{Additional images}
\begin{figure*}[h]
\begin{subfigure}{1.0\linewidth}
  \centering
  \rotatebox{90}{\begin{minipage}{.0558\linewidth}\center\tiny image\end{minipage}}
  \incpr{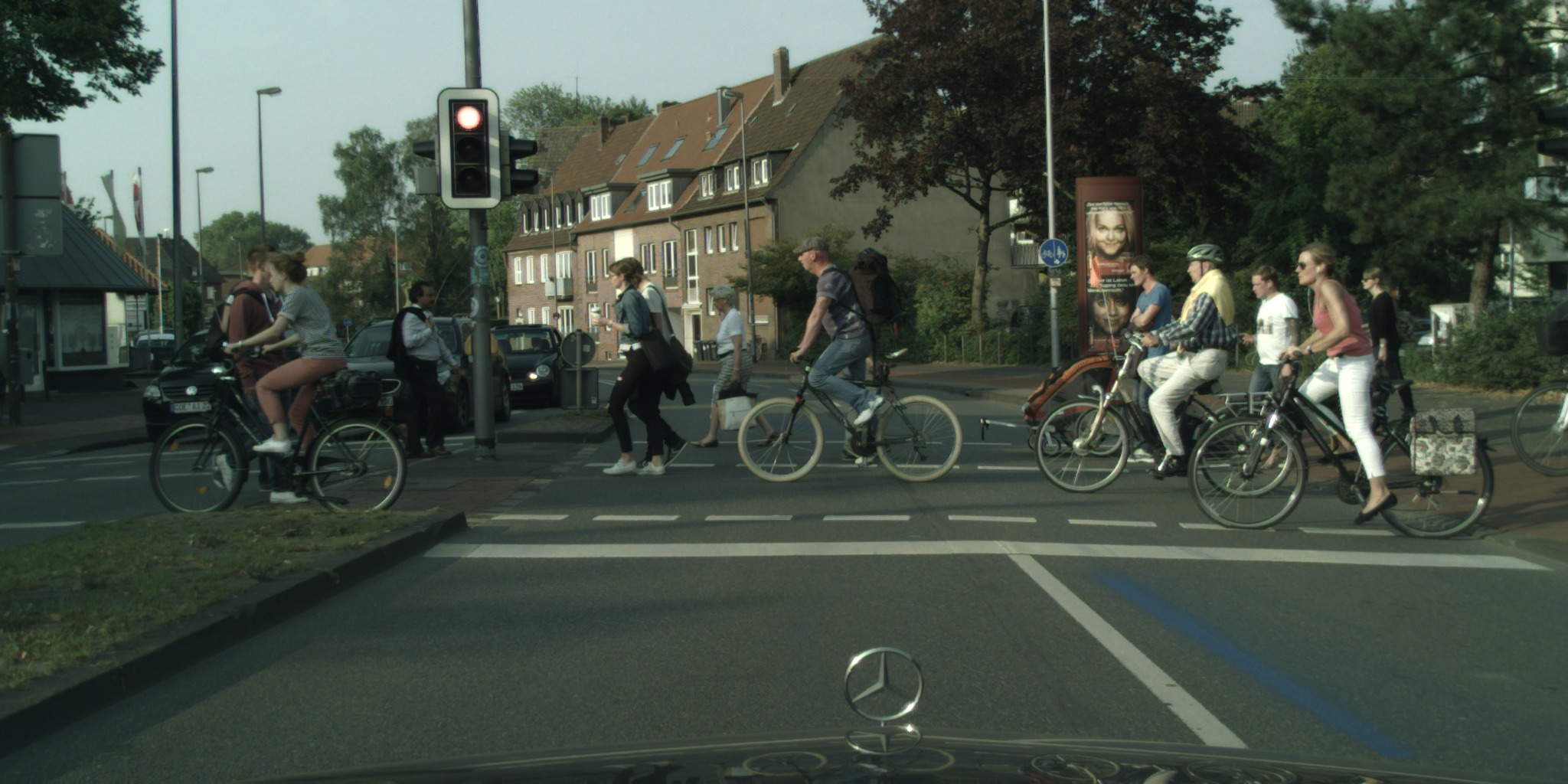}
  \incpr{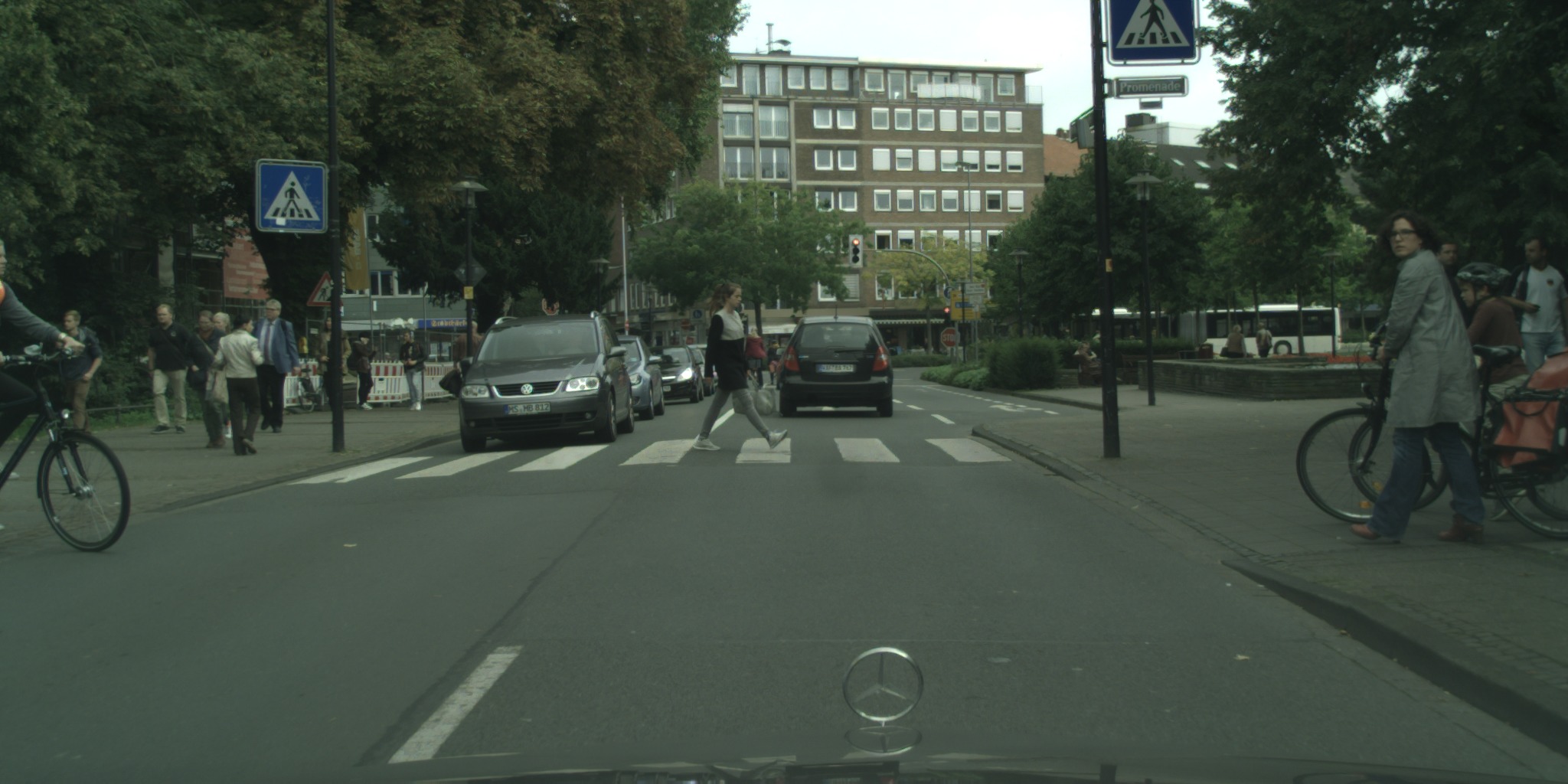}
  \incpr{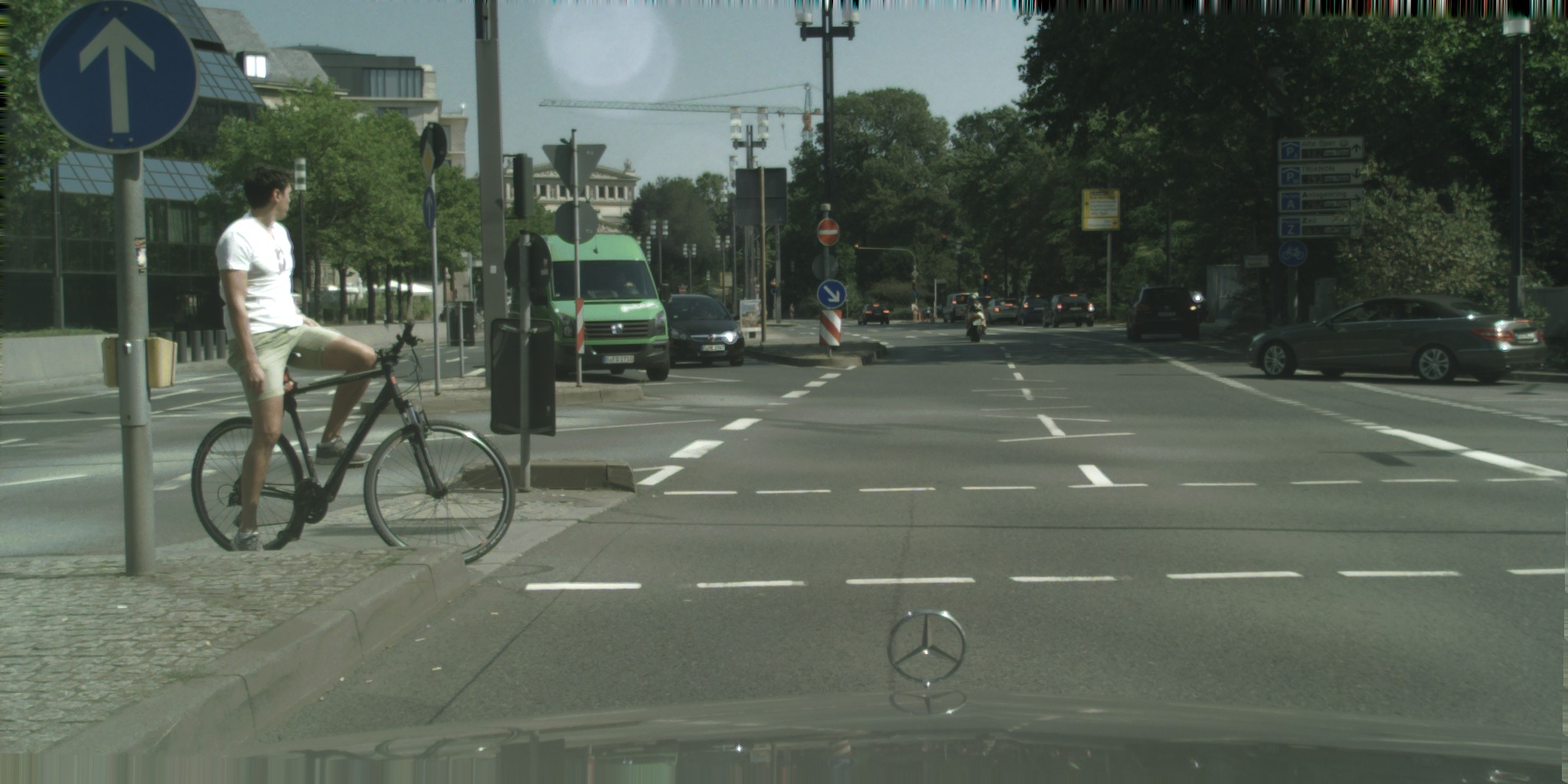}
  \incpr{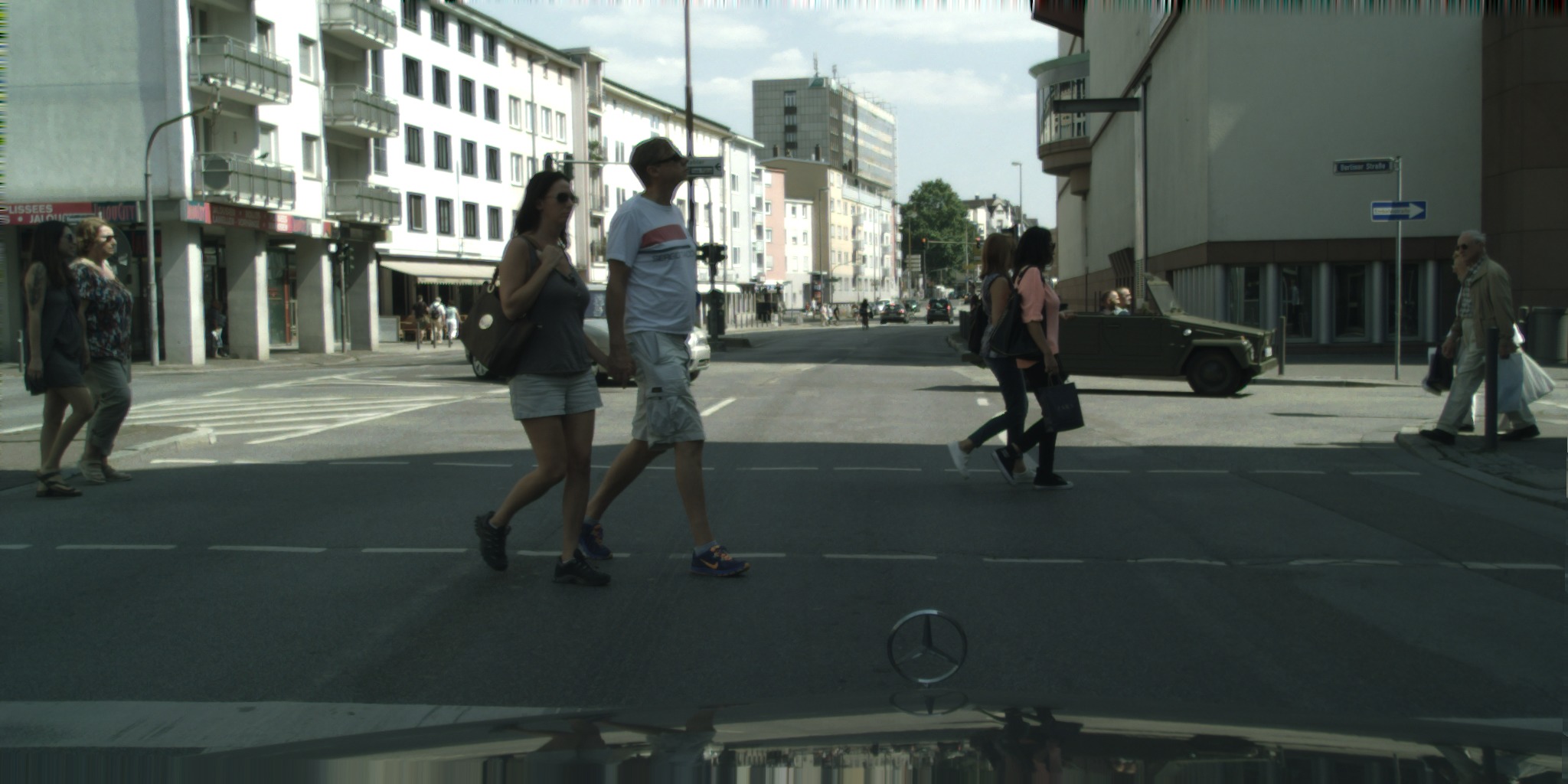}
\end{subfigure}\\[1pt]
\begin{subfigure}{1.0\linewidth}
  \centering
  \rotatebox{90}{\begin{minipage}{.0518\linewidth}\center\tiny groundtruth\end{minipage}}
  \incprgt{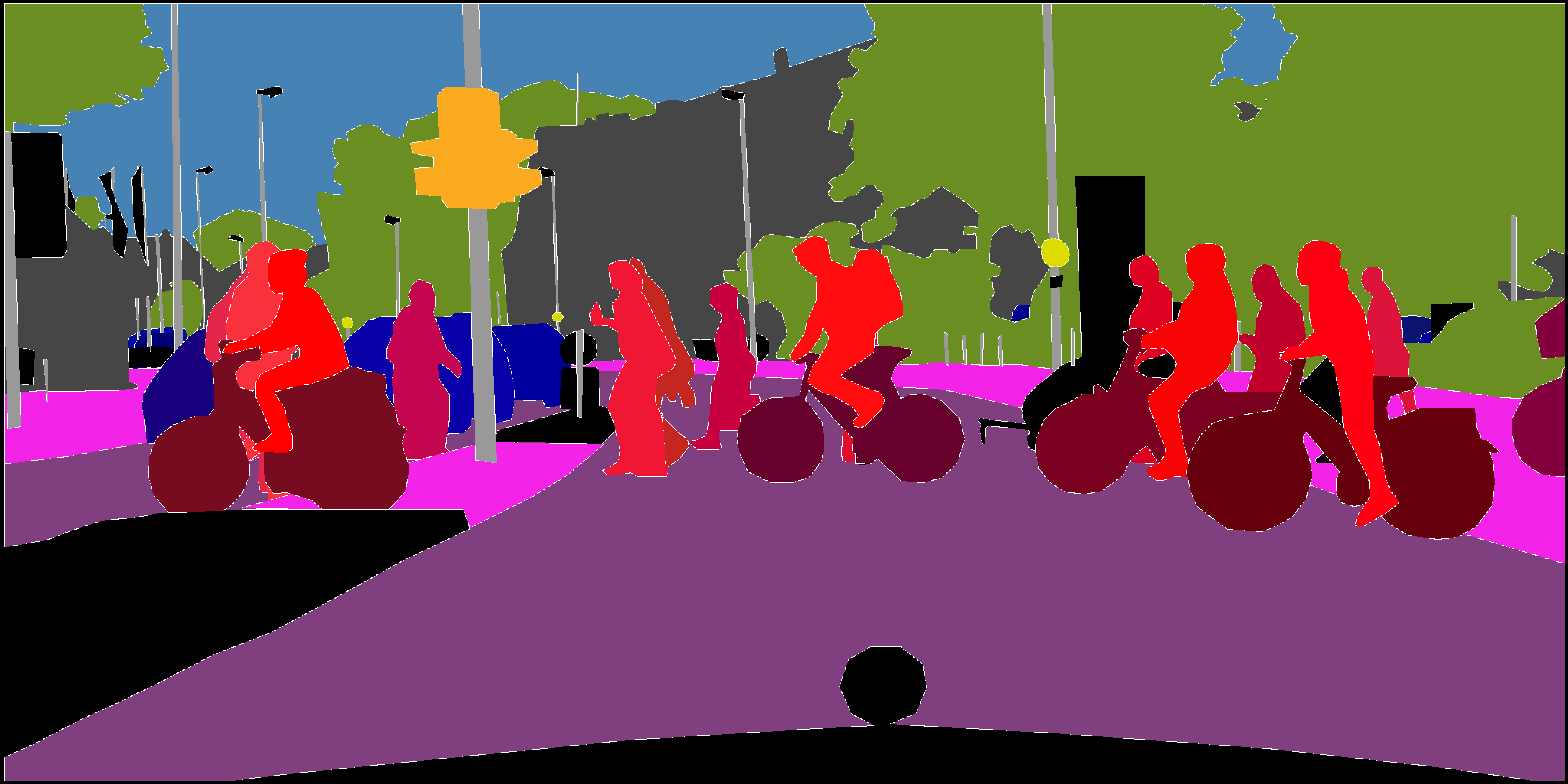}
  \incprgt{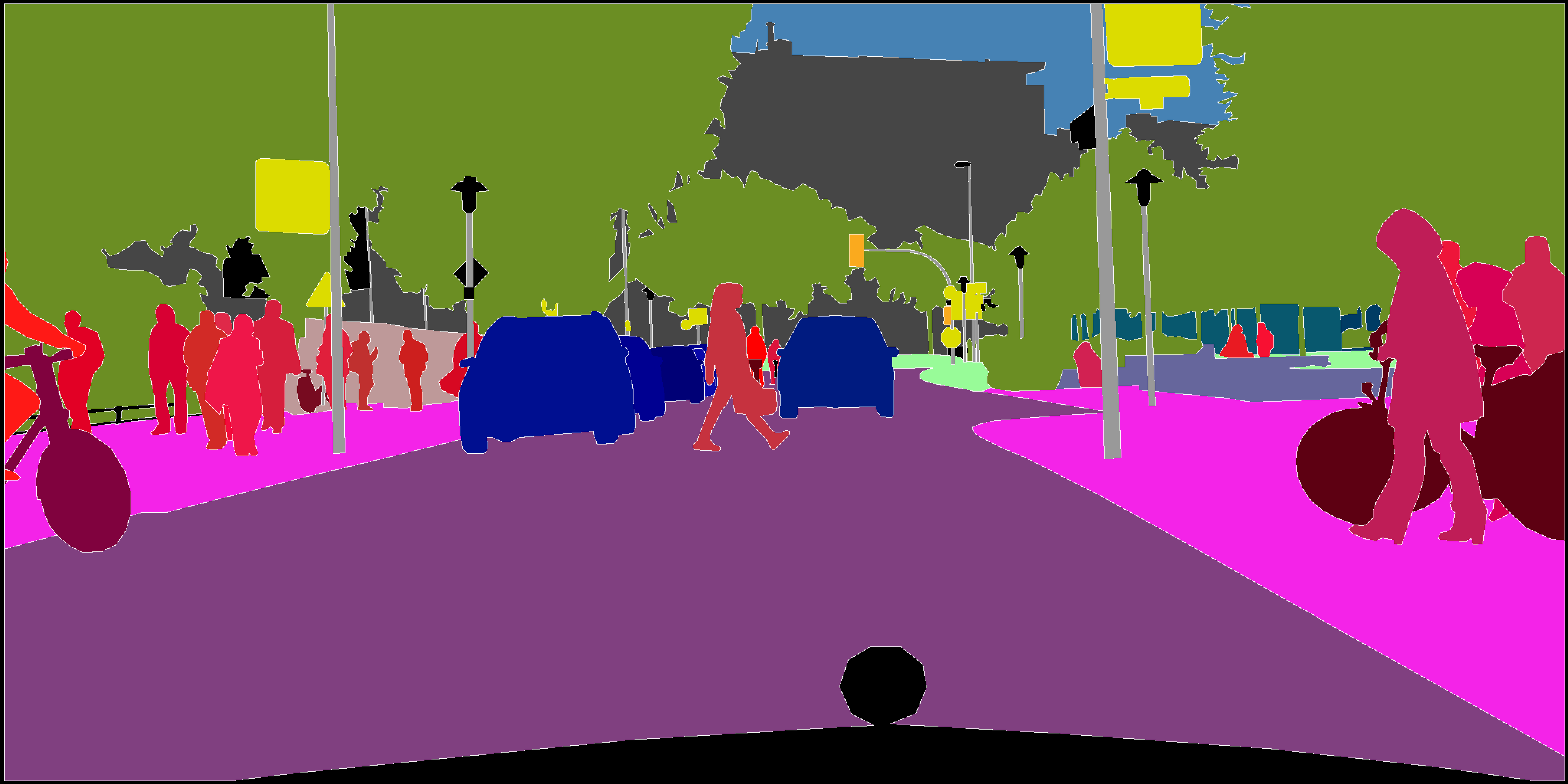}
  \incprgt{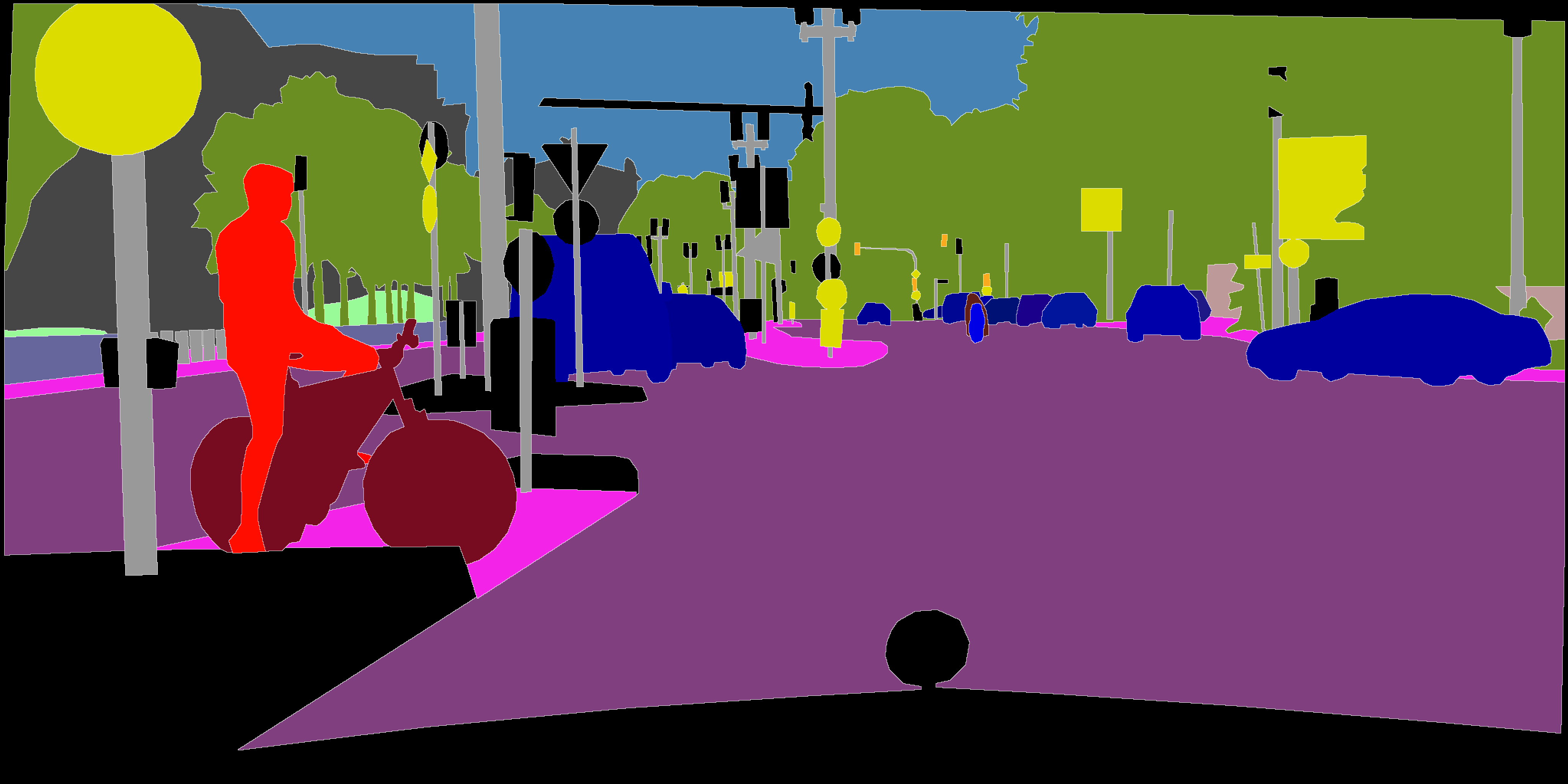}
  \incprgt{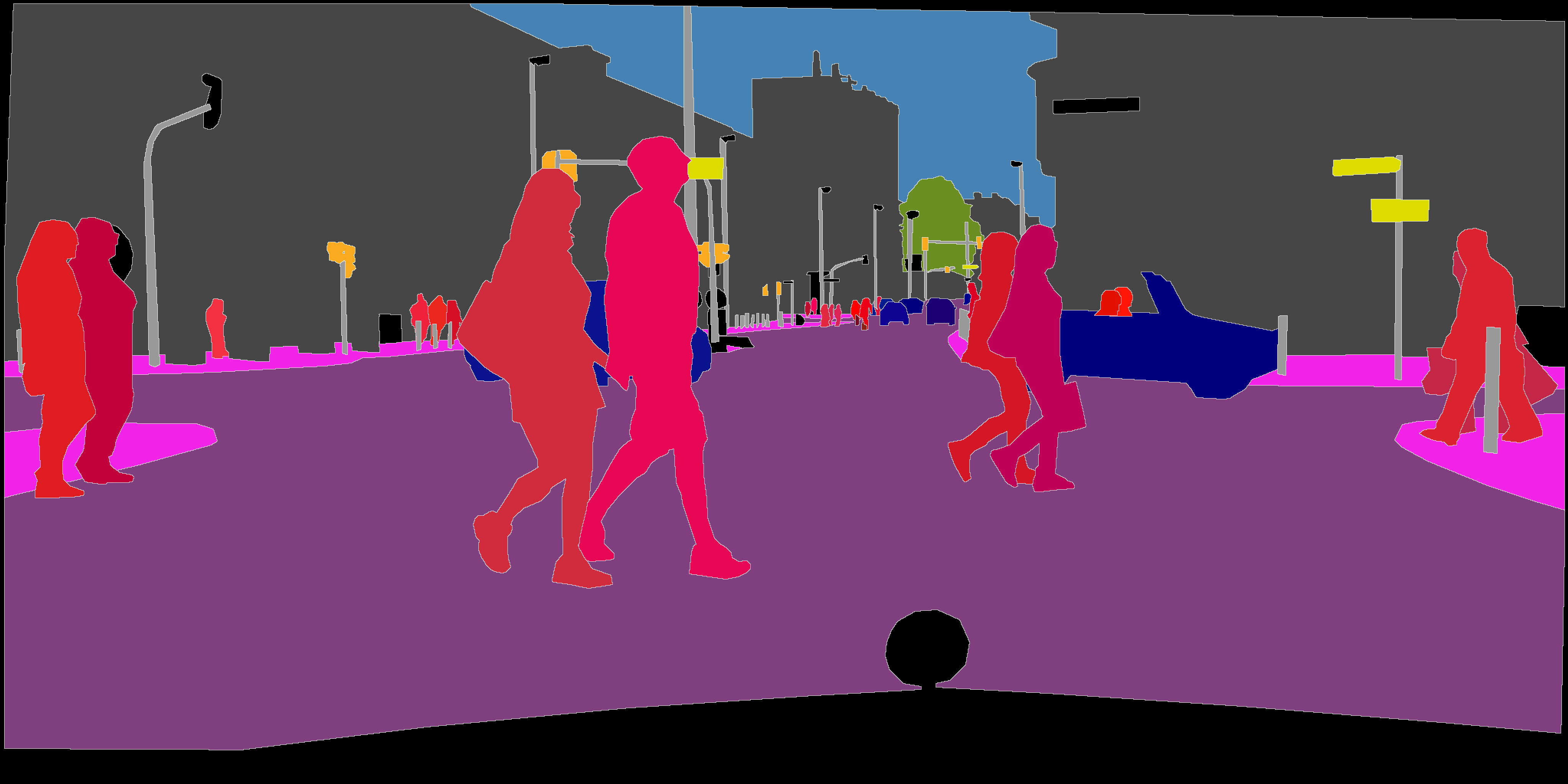}
\end{subfigure}\\[1pt]
\begin{subfigure}{1.0\linewidth}
  \centering
  \rotatebox{90}{\begin{minipage}{.0518\linewidth}\center\tiny prediction\end{minipage}}
  \incprpred{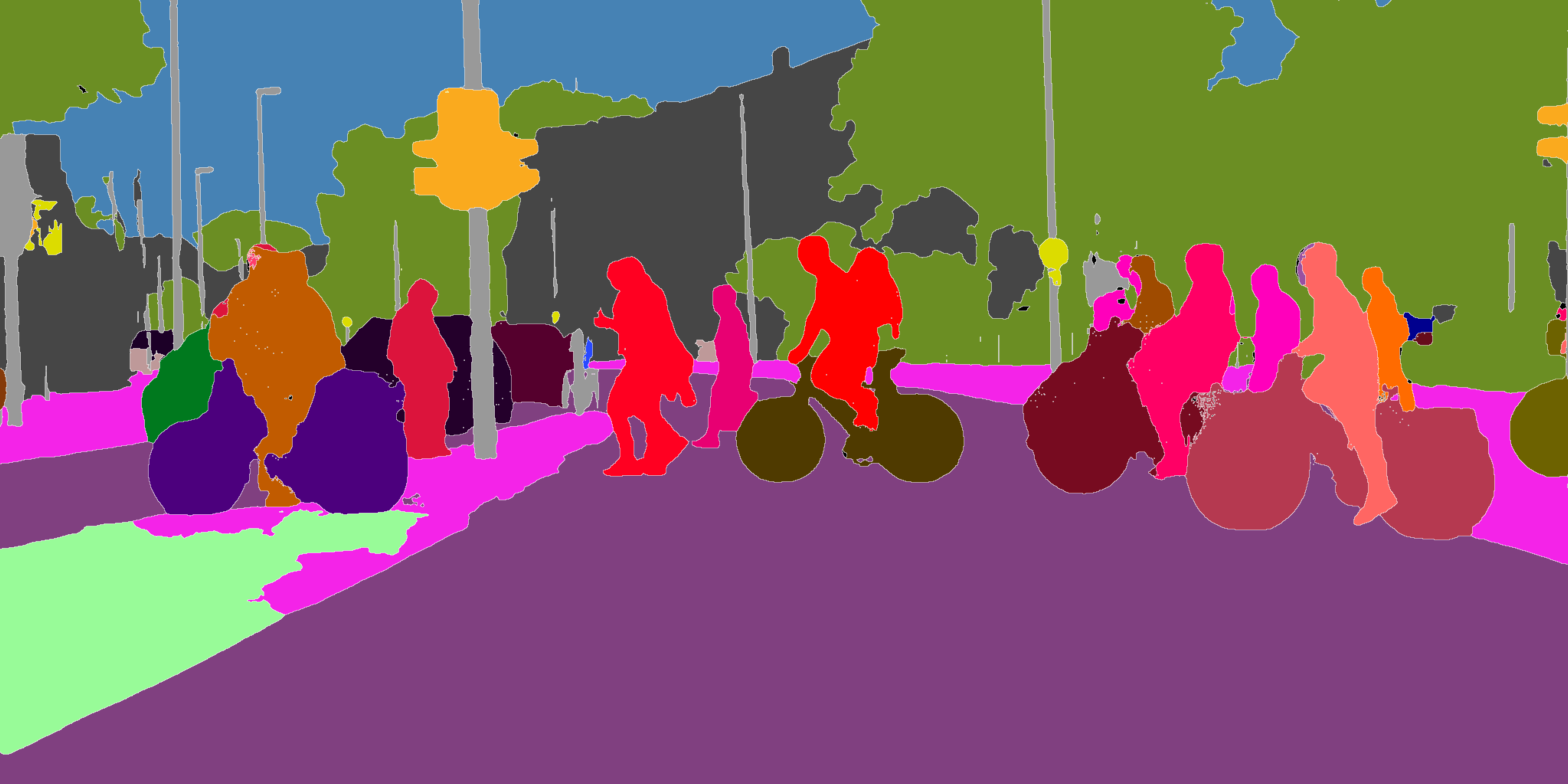}
  \incprpred{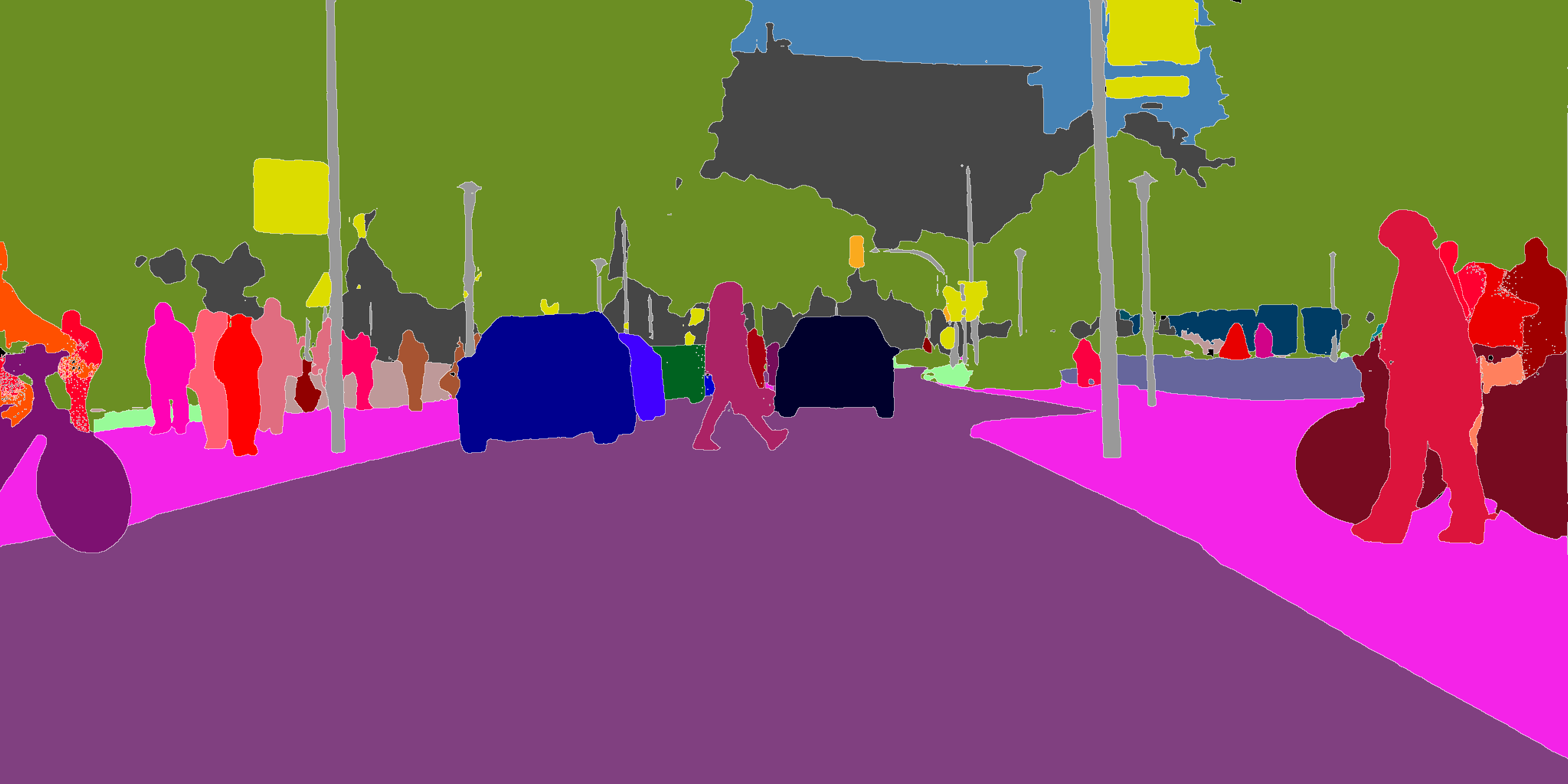}
  \incprpred{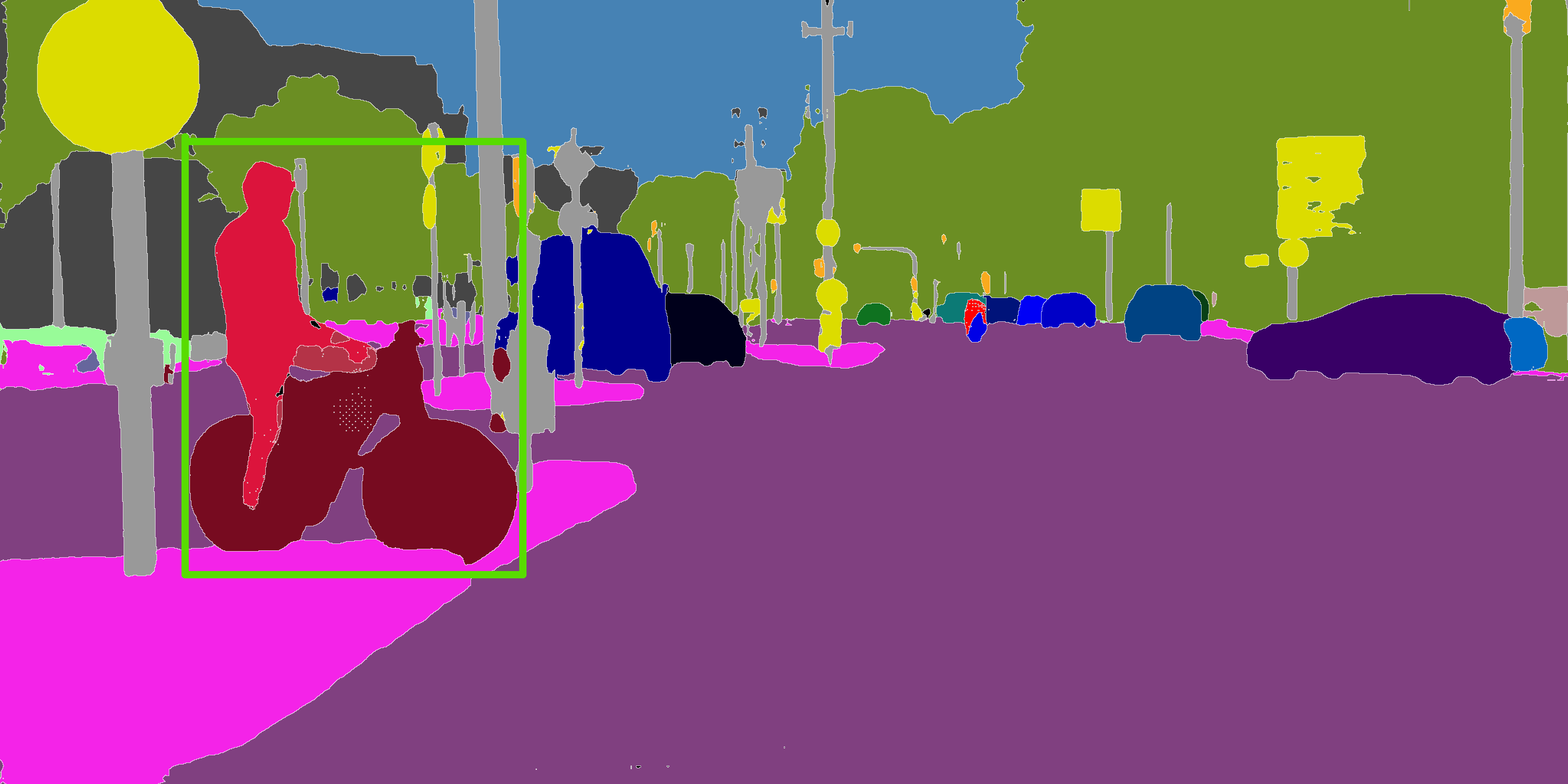}
  \incprpred{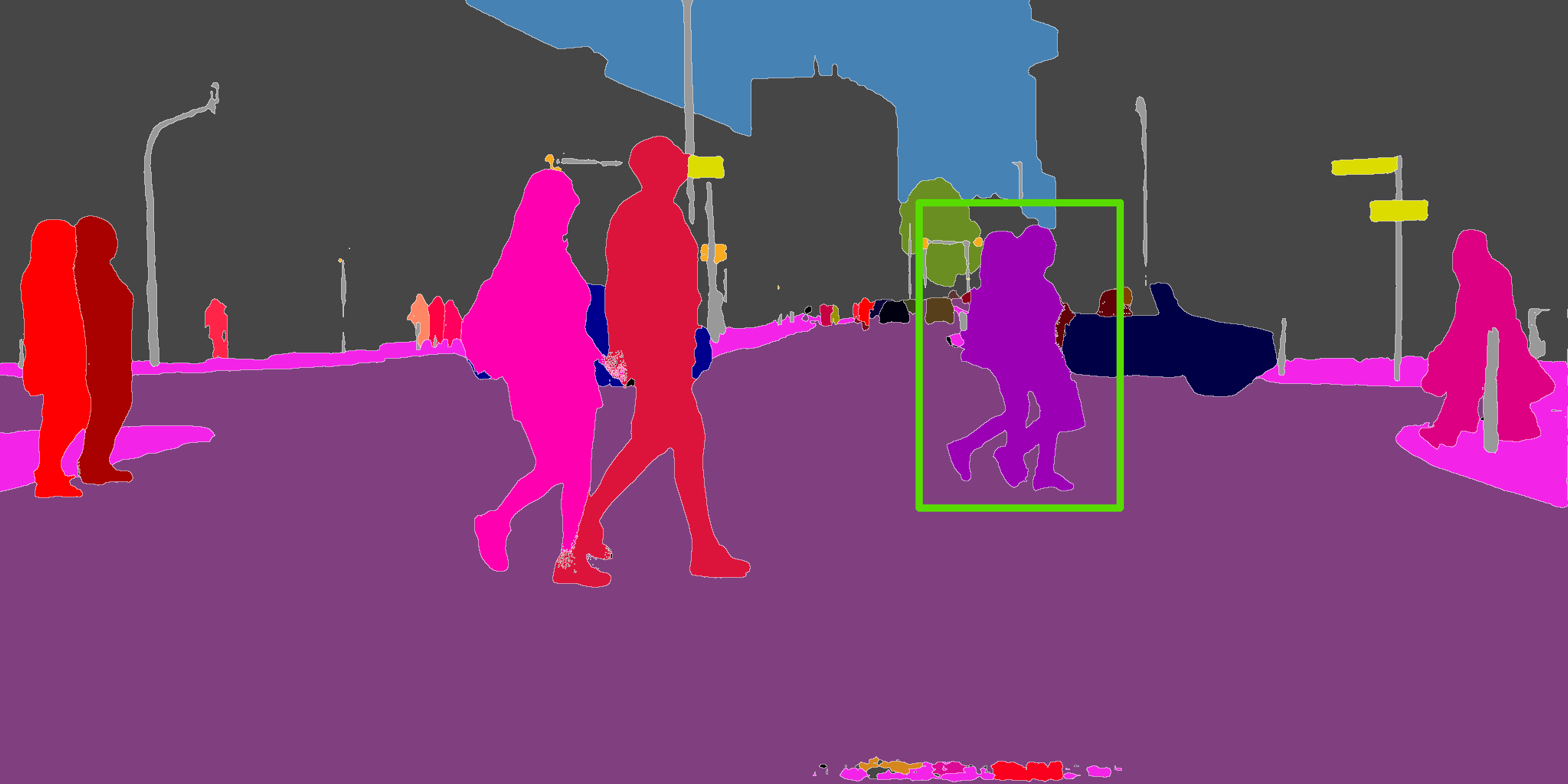}
\end{subfigure}
\vspace{2mm}
\caption{Further examples panoptic results on Cityscapes using using semantic unaries (Deeplab 3+ network) and affinities derived from Mask-RCNN foreground probability. Prediction errors are highlighted in green.}
\label{fig:examples_apdx}
\vspace{-2mm}
\end{figure*}

\section{Scaling Behavior}\label{sec:scaling}
\begin{figure*}[h]
\includegraphics[width=\linewidth]{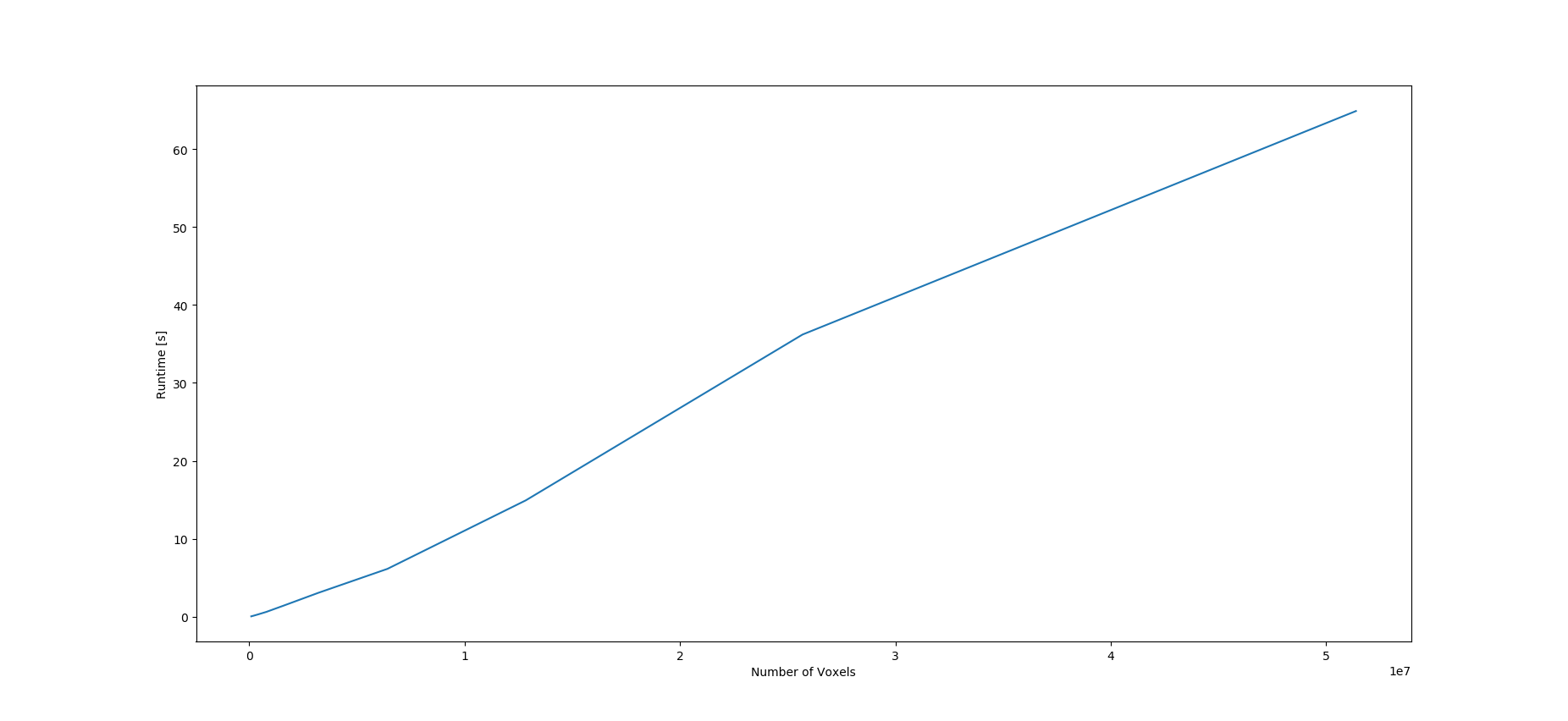}
\caption{Runtime scaling of the SMW. We evaluate the runtime of the SMW for different volume sizes of the 3D Sponge dataset. We find an almost linear relation between runtime and number of voxels.}
\label{fig:rt_apdx}
\vspace{-2mm}
\end{figure*}